%% file: main.tex
\documentclass{article}

\PassOptionsToPackage{numbers,compress}{natbib}
\usepackage[final]{neurips_2026}

\workshoptitle{}

\makeatletter
\renewcommand{\@noticestring}{%
  Submitted to \@neuripsordinal\ Conference on Neural Information Processing Systems
  (NeurIPS \@neuripsyear).%
}
\makeatother
\usepackage[utf8]{inputenc}
\usepackage[T1]{fontenc}

\usepackage{amsmath,amssymb,mathtools,amsthm}
\usepackage{graphicx}
\usepackage{booktabs,multirow}
\usepackage{array}
\newcolumntype{L}[1]{>{\raggedright\arraybackslash\hspace{0pt}}p{#1}}

\usepackage{xcolor}
\usepackage{tikz}
\usetikzlibrary{positioning,arrows.meta,fit,backgrounds,calc,shapes.geometric,shapes.symbols,decorations.pathmorphing}
\input{icons}
\usepackage{pifont}

\usepackage{enumitem}
\usepackage{caption}
\usepackage{subcaption}
\usepackage{algorithm}
\usepackage{algpseudocode}

\usepackage[
  colorlinks=true,
  linkcolor=blue!55!black,
  citecolor=blue!55!black,
  urlcolor=blue!55!black
]{hyperref}
\ifdefined\pdfinfoomitdate \pdfinfoomitdate=1 \fi
\ifdefined\pdftrailerid \pdftrailerid{} \fi

\hypersetup{
  pdftitle={},
  pdfauthor={},
  pdfsubject={},
  pdfkeywords={meta-agents, architecture search, non-stationarity, conservative architecture updates},
  pdfcreator={},
  pdfproducer={}
}

\usepackage{microtype}

\newtheorem{proposition}{Proposition}
\newtheorem{definition}{Definition}

\newcommand{\cmas}{\textsc{Cmas}}
\newcommand{\bench}{\textsc{MetaInvest-Bench}}

\newcommand{\Jc}{\tilde J} 
\newcommand{\Score}{\mathrm{Score}} 
\title{Conservative Meta-Agent Search:\\
 Architecture Adaptation under Non-Stationarity}

\author{Ali Atiah Alzahrani$^{\dagger}$}

\begin{document}
\maketitle

\begingroup
\renewcommand{\thefootnote}{$\dagger$}%
\footnotetext{Corresponding author: \texttt{aliatiah100@gmail.com}. The views
expressed are those of the author and do not reflect the views of any other
individual or entity. This material is for research purposes only and does not
constitute investment advice.}%
\endgroup

\begin{abstract}
\noindent Automated design of agentic systems is typically evaluated on static task suites with immediate scalar rewards. Deployed systems face persistent shift, delayed dependent feedback, correlated component failures, and costly architecture changes. We formalize meta-agent architecture search as \emph{sequential incumbent replacement} with switching costs and hard risk constraints, and introduce Conservative Meta-Agent Search (\cmas). Throughout, \emph{meta-agent} denotes the entire outer loop: proposer, compiler, evaluator, gate, and rollback controller. \cmas{} retrieves archived architectures and proposes typed graph edits, then scores each against the incumbent by paired shadow replay under common random numbers. Component credit is interventional. Promotion requires an \emph{empirically calibrated, variance-penalized simultaneous score} above a pre-specified margin, and is followed by a simulated canary stage and pre-specified rollback. Backtested return cannot certify meta-level skill, so we introduce \bench{}, a regime-switching simulator whose specified worker competence makes a surrogate-optimal architecture computable, with leakage-controlled replay, adversarial governance tracks, and five reusable negative controls. Financial streams give an instrumented, risk-sensitive test of meta-level generalization; we claim no alpha. Across 96 streams, \cmas{} cuts normalized mean surrogate dynamic-oracle regret from 0.116 to 0.104 and raises held-out utility from 0.44 to 0.48 over the strongest adaptive common-budget reimplementation; when extra workers add no value, it contracts to near-single-agent teams. Its \emph{deployment} inference cost is ${\sim}6\%$ lower, though its search-inclusive token cost is the highest of the methods compared. The observed per-update false-promotion rate is 3.4\% against a nominal 5\% target, and 4.8\% in a sealed rerun. The statistical ingredients are standard. The calibration is empirical: it concerns an update-level replay statistic on an author-designed benchmark, and guarantees neither full-utility improvement, nor absence of simulated canary-stage harm, nor sequence-level error control.
\end{abstract}


\section{Introduction}
\label{sec:intro}

Meta-agents act on agentic systems. They design harnesses, optimize compound workflows, allocate resources, and supervise workers. An LLM system's performance depends on weights, but also on prompts, tools, memory, communication, verification, and control flow~\cite{gepa,adas,metaharness}, and a growing literature searches code-defined workflows~\cite{aflow}, modular components~\cite{agentsquare}, communication graphs~\cite{gptswarm,gdesigner}, and query-dependent architectures~\cite{automaas,maas}. Yet most evaluations remain \emph{static}: candidates receive immediate scores on exchangeable tasks, and the best design is deployed without modeling persistent change or the risk of a bad update.\looseness=-1{}

\begin{figure}[t]
\centering
\resizebox{0.93\textwidth}{!}{\input{fig_motivation}}
\caption{\textbf{Illustrative schematic of architecture as a sequential decision variable; annotations are conceptual and do not report experimental results.} A frozen workflow may miss persistent shifts; a fixed committee stays expensive in every regime; \cmas{} expands, rewires, contracts, retains, or rolls back under fixed external controls, after reversible calibrated promotion.\looseness=-1{}}
\label{fig:motivation}
\end{figure}
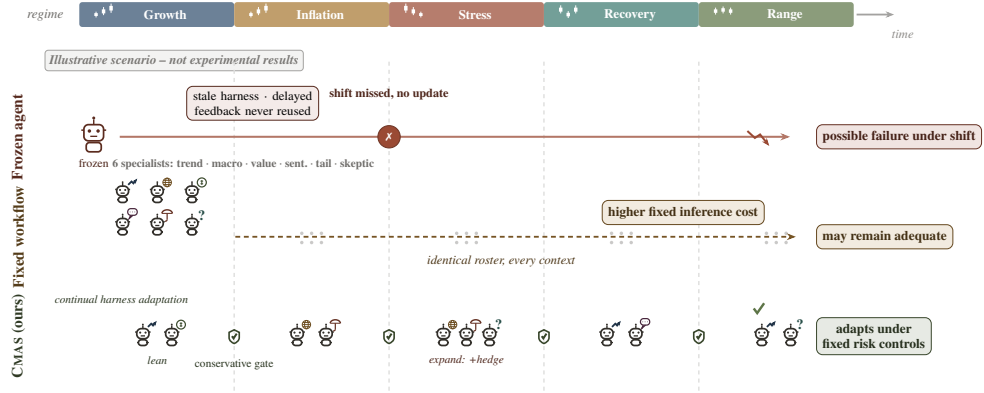

Architecture is therefore a \emph{decision variable} that should track the regime (Figure~\ref{fig:motivation}). Investment systems expose non-stationarity, delayed and dependent feedback, correlated component errors, switching costs, and tail-risk constraints. Finance is useful here as an instrumented, risk-sensitive test of meta-level generalization and conservative update evaluation. Backtested return is not intrinsically persuasive~\cite{profitmirage,lookahead,livebench}. What is missing is (i) a controlled benchmark whose surrogate-optimal worker architecture is computable and changes over time, and (ii) a method that \emph{screens} candidate updates under noisy, delayed feedback~\cite{finmem,tradingagents,atlas,selfdriving}. An update whose simulator-evaluated mean improvement is at most the promotion margin is a \emph{false promotion}, the meta-level analogue of a false discovery and a near-neutral change; Table~\ref{tab:defs} separates it from the outcomes that do make the system worse.\looseness=-1{}

Generated multi-agent systems can also be costlier yet weaker than strong single-agent baselines~\cite{illusion}; a credible meta-agent must learn both when to assemble specialists and when to collapse to a simpler system. We ask whether a meta-agent can adapt the composition, communication graph, and compute allocation of a compound agentic system under regime shifts while empirically limiting false promotion, architecture bloat, and operational risk.\looseness=-1{}

We answer with \cmas{} (Figure~\ref{fig:overview}). The \emph{inner loop} executes a bounded agent architecture behind a deterministic constraint and authorization layer. The \emph{outer loop} observes the pre-decision regime descriptor $z_k$ and a trace archive, proposes typed edits, evaluates candidates against the incumbent by paired shadow replay, assigns counterfactual credit, and promotes only when the empirically calibrated simultaneous score clears the margin. Promotion enters a simulated canary stage and is reversible. Every method sees the same pre-decision descriptor $z_k$, available before the decisions it informs, while the latent simulator regime $s_t$ is evaluator-only. Throughout, \emph{meta-agent} denotes this entire outer loop (proposer, compiler, evaluator, credit estimator, gate, and rollback controller), not the proposal model alone. It is a \emph{hybrid}: generation is LLM-based and archive-conditioned, while the compiler, risk projection, authorization layer, gate, and rollback rules are deterministic and outside anything the LLM can edit (Figure~\ref{fig:overview}, locked box).\looseness=-1{}

Three contributions follow. (1)~We formulate agent-architecture search as sequential \emph{incumbent replacement} over typed architectures, with a structural distance (Eq.~\eqref{eq:dist}) as switching cost, delayed dependent utility (Eq.~\eqref{eq:utility}), hard eligibility constraints, an explicit false-promotion target, and meta-level generalization (Definition~\ref{def:metagen}) as the scored quantity. (2)~\cmas{} couples archive-conditioned typed edits, paired candidate--incumbent replay under common random numbers, interventional Shapley-style~\cite{shapley,shap} credit fed back to the proposer, and a conservative promotion-and-rollback gate. The statistical ingredients are standard; the contribution is their integration. Proposition~\ref{prop:fp} states only the \emph{idealized} conditions under which such a gate would control family-wise false promotion, and Section~\ref{sec:method} scopes the departures. (3)~\bench's controlled track makes surrogate-oracle regret, selection accuracy, and credit fidelity computable, and five reusable controls (clone-collapse, shuffled-regime, gate-calibration, proposer-value, Goodhart-drift) measure whether adaptation happens for the intended reason, exposing reward gaming under a return-only evaluator (RQ7).\looseness=-1{}

The main empirical finding is a \emph{promotion-calibration separation}. Holding the evaluator and gate fixed, false promotion stays at 3.2--3.5\% across proposers while utility tracks proposer quality (Table~\ref{tab:ablate2}).\looseness=-1{}


\section{Related work and positioning}
\label{sec:related}

\textbf{Automated design and optimization of agent systems.} ADAS, AgentSquare/AFlow, MaAS/AutoMaAS, and GEPA search executable programs, workflows, architectures, or prompts~\cite{adas,agentsquare,aflow,maas,automaas,gepa}. GPTSwarm and G-Designer optimize communication graphs~\cite{gptswarm,gdesigner}. Adaptive Auto-Harness directly studies sustained harness construction and routing on open-ended, heterogeneous, non-stationary streams~\cite{adaptiveharness}. HarnessAudit evaluates trajectory-level boundary and information-flow compliance, while SafeHarness integrates verification, privilege control, degradation, and rollback across the agent lifecycle~\cite{harnessaudit,safeharness}. What \cmas{} makes first-class is repeated candidate--incumbent replacement, paired replay on shared randomness, update-level promotion calibration, switching cost, and a reversible simulated canary stage. The closest systems route at solve time and leave the deployed incumbent in place~\cite{adaptiveharness,maas}, gate on candidate quality without a paired incumbent comparison or rollback~\cite{gatedqd,harbor}, or audit and secure execution without an update-level false-promotion estimand~\cite{harnessaudit,safeharness}; full comparisons in Appendix~\ref{app:related}.\looseness=-1{}

\textbf{Safe improvement, verification, and evaluation validity.} Four literatures supply ingredients (Appendix~\ref{app:related}): safe policy improvement and conservative bandits give the gate its form~\cite{hcpi,spibb,consbandit}; non-stationary switching-cost learning the regret metric~\cite{besbes,herbster}; safety and verification work the threat model~\cite{mama,verifmeta,machallenge}; off-policy evaluation~\cite{ope} motivates paired replay. New here are a typed action space with a distance (Eq.~\eqref{eq:dist}), adaptively LLM-generated pools (so the guarantee is pool-conditional), and interventional trace-level credit. Because financial evaluation is vulnerable to leakage~\cite{profitmirage,lookahead,livebench,memctrl,finmem,tradingagents,atlas,expertteams,selfdriving}, the simulator is primary and market replay provides external validity only.\looseness=-1{}


\section{Problem: sequential meta-agent design under non-stationarity}
\label{sec:problem}

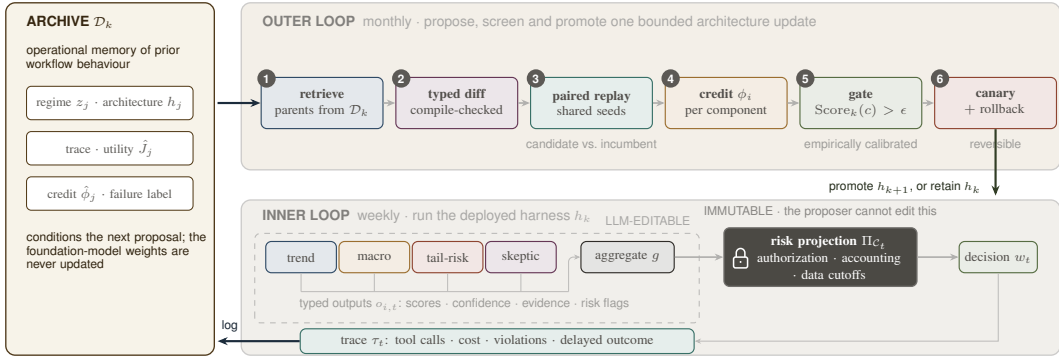
\begin{figure}[t]
\centering
\resizebox{1\textwidth}{!}{\input{fig_overview}}
\caption{\textbf{\cmas{} as a governed adaptation loop for a multi-agent investment workflow.} The risk projection, authorization, accounting and data cutoffs sit outside the LLM-editable boundary. \emph{Left:} the archive $\mathcal{D}_k$ is operational memory of prior workflow behavior, storing regimes, architectures, traces, utility and credit estimates, and failure labels; the foundation-model weights are never updated. \emph{Bottom (inner loop, weekly):} incumbent $h_k$ runs typed specialists whose outputs $o_{i,t}$ are aggregated by $g$, after which the immutable projection $\Pi_{\mathcal{C}_t}$ enforces the financial control boundary and the trace $\tau_t$ is logged back to the archive. \emph{Top (outer loop, monthly):} \ding{182} retrieve parents by regime similarity and novelty; \ding{183} propose a compile-checked, auditable typed diff with hypothesis; \ding{184} paired shadow replay of candidate against incumbent under common random numbers, plus red-teaming; \ding{185} interventional credit $\phi_i$ fed back; \ding{186} promote only if the empirically calibrated simultaneous score exceeds $\epsilon$; \ding{187} simulated canary stage with rollback, after which the promoted $h_{k+1}$ (or the retained $h_k$) returns to the inner loop.}

\label{fig:overview}
\end{figure}

\paragraph{Inner decision system.}
At time $t$ an information packet $x_t$ is $\mathcal{F}_t$-measurable; architecture $h$ produces an unconstrained proposal $\tilde w_t(h)$ and trace $\tau_t(h)$. A deterministic projection maps it to the feasible set, $w_t(h) = \Pi_{\mathcal{C}_t}(\tilde w_t(h))$, with $\mathcal{C}_t$ encoding leverage, concentration, turnover, volatility, and restricted-asset constraints. The LLM system cannot modify $\Pi_{\mathcal{C}_t}$, accounting, authorization, or data cutoffs (Figure~\ref{fig:overview}, locked box). Net return is $R^{\mathrm{net}}_{t+1}(h) = w_t(h)^{\!\top} r_{t+1} - \kappa_t\lVert w_t(h)-w_{t-1}(h)\rVert_1$; during paired replay $w_{t-1}(h)$ is the candidate's \emph{own} counterfactual weights, so each pays its own path's turnover. Because an active projection can mask proposal differences, we fix its form throughout: Euclidean projection onto the constraint polytope via a QP solver.\looseness=-1{}

\paragraph{Architectures, distance, and meta-level generalization.}
An executable architecture is a typed object $h = (V, E, \rho, \mu, b, g, v) \in \mathcal{H}$: worker set $V$, directed acyclic communication graph $E$, typed roles and prompts $\rho$, model assignments $\mu$, compute/tool budgets $b$, aggregation rule $g$, verification nodes $v$. We equip $\mathcal{H}$ with a distance
\begin{equation}
\begin{aligned}
d(h,h') ={}& \omega_V d_{\mathrm{Jac}}(V,V') + \omega_E d_{\mathrm{GED}}(E,E') \\
&+ \omega_b d_{\mathrm{budget}}(b,b')
+ \omega_\rho d_{\mathrm{typed}}\!\big((\rho,\mu,g,v),(\rho',\mu',g',v')\big).
\end{aligned}
\label{eq:dist}
\end{equation}
used both as a switching cost and as a drift diagnostic. All four terms are normalized to $[0,1]$, so no architecture field changes at zero switching or novelty cost. With $\le$10 edges (Appendix~\ref{app:space}) $d_{\mathrm{GED}}$ is exact; weights are fixed in advance (Table~\ref{tab:hparams}) and act through $\lambda_{\mathrm{switch}}$. At outer-loop time $k$, the meta-agent sees the observable pre-decision regime descriptor $z_k$, the incumbent $h_k$, and an archive $\mathcal{D}_k$ of traces, credits, and outcomes; it proposes a candidate pool $\mathcal{H}_k \subset \mathcal{H}$.\looseness=-1{}

\begin{definition}[Meta-level generalization]
\label{def:metagen}
A meta-policy $\pi$ \emph{$(\varepsilon_{\mathrm{gen}},\varepsilon_{\mathrm{safe}})$-generalizes} over a held-out factor family $F$ (tasks, regime mixtures, worker libraries, model backbones) if expected held-out utility is within $\varepsilon_{\mathrm{gen}}$ of in-distribution utility and each held-out violation or false-promotion rate is no more than $\varepsilon_{\mathrm{safe}}$ worse; the strict version sets $\varepsilon_{\mathrm{safe}}=0$. We report achieved deltas rather than asserting a threshold (Table~\ref{tab:transfer}): the strict clause is \emph{not} met, since held-out false promotion rises 0.7\,pp (RQ3).
\end{definition}

Constrained utility over shadow window $W_k$ scalarizes soft preferences; hard constraints stay eligibility filters. A candidate failing authorization, evidence, leakage, or risk checks is ineligible regardless of utility. For candidate $h$ and the sequence's previous architecture $h^{-}$,
\begin{equation}
\begin{aligned}
J_k(h;h^{-}) \;=\;& \mathrm{CEQ}_{\gamma}\!\big(R^{\mathrm{net}}(h)\big) \;-\; \lambda_{\mathrm{cvar}}\,\mathrm{CVaR}_{.95}\!\big(-R^{\mathrm{net}}(h)\big) \;-\; \lambda_{\mathrm{dd}}\,\mathrm{MDD}(h) \\
&-\; \lambda_{\mathrm{cost}}\,\mathrm{Cost}(h) \;-\; \lambda_{\mathrm{turn}}\,\mathrm{Turnover}(h) \;-\; \lambda_{\mathrm{switch}}\, d(h,h^{-}) \;-\; \lambda_{\mathrm{viol}}\,\mathrm{Viol}(h).
\end{aligned}
\label{eq:utility}
\end{equation}
$\mathrm{CEQ}_{\gamma}(R)=\mathbb{E}[R]-\tfrac{\gamma}{2}\mathrm{Var}(R)$ is the certainty equivalent, $\mathrm{CVaR}_{.95}$ the mean worst 5\%, $\mathrm{MDD}$ maximum drawdown; penalties are nonnegative and normalized by frozen training statistics (Appendix~\ref{app:repro}). $\mathrm{Viol}$ counts only \emph{soft} violations (hard authorization and risk violations are eligibility filters), and ``OOD utility'' is held-out $J_k$ under the same frozen constants. The gate tests only the additive return/cost/turnover/switch statistic $\Jc$, which omits the path-risk terms, so a candidate can clear the gate yet reduce full $J_k$ within eligibility. A \emph{false promotion} is distinct from both a \emph{negative-$\Jc$ promotion} and a \emph{full-$J$ reduction} (Table~\ref{tab:defs}). Let $h^{\star}_{1:K}$ denote the computable surrogate dynamic-oracle sequence that maximizes $\sum_k J_k(h^{\star}_k;h^{\star}_{k-1})$ over the enumerated feasible surrogate space. The reported normalized mean surrogate dynamic-oracle regret is $K^{-1}\sum_k\!\left[J_k(h^{\star}_k;h^{\star}_{k-1})-J_k(h_k;h_{k-1})\right]$; each sequence pays switching cost against its own predecessor.\looseness=-1{}

The promotion gate uses the following explicitly additive block statistic, with every term normalized by the same frozen training constants used in Eq.~\eqref{eq:utility}:
\begin{equation}
\begin{aligned}
\Jc_{k,b}(h;h^{-}) ={}& \overline{R}^{\mathrm{net}}_{k,b}(h)
-\lambda_{\mathrm{cost}}\mathrm{Cost}_{k,b}(h)
-\lambda_{\mathrm{turn}}\mathrm{Turnover}_{k,b}(h)\\
&-\lambda_{\mathrm{switch}}d(h,h^{-})
-\lambda_{\mathrm{viol}}\mathrm{Viol}_{k,b}(h).
\end{aligned}
\label{eq:certified}
\end{equation}
The CEQ variance term, CVaR, and MDD are excluded from $\Jc$ as non-additive path functionals; they act through full-$J$ reporting and hard eligibility filters. A candidate pays $d(c,h_k)$, while the incumbent's own comparison $\Jc_{k,b}(h_k;h_k)$ pays zero incremental switching cost. False-promotion labels concern the held-out mean of this paired statistic. They say nothing about full utility or downstream harm.\looseness=-1{}

\section{\cmas: conservative meta-agent search}
\label{sec:method}

\paragraph{Typed worker and architecture interfaces.}
Workers are bounded specialists, each returning a typed tuple $o_{i,t} = (s_{i,t}, c_{i,t}, u_{i,t}, e_{i,t}, q_{i,t})$: asset scores, calibrated confidence, uncertainty, evidence identifiers, risk flags. The library holds trend, macro, value/carry, sentiment, liquidity/tail-risk, and skeptic agents. Before execution, an architecture \emph{compiler} validates schemas, acyclicity, permissions, budgets, and evidence contracts, rejecting invalid diffs with machine-readable errors the proposer can learn from.\looseness=-1{}

\paragraph{Archive-conditioned candidate generation.}
The archive stores tuples $(z_j, h_j, \tau_j, \hat J_j, \hat\phi_j, f_j)$, where $\hat\phi_j$ are component credits and $f_j$ failure labels. Parents are retrieved by
\begin{equation}
S_j \;=\; \beta_z\,\mathrm{sim}(z_k, z_j) \;+\; \beta_J\, \hat J_j \;-\; \beta_u\, \hat\sigma_j \;+\; \beta_d\, \mathrm{novelty}(h_j),
\label{eq:retrieve}
\end{equation}
with Gaussian regime kernel $\mathrm{sim}(z_k,z_j)=\exp(-\|z_k-z_j\|^2/2\ell^2)$ and $\mathrm{novelty}(h_j)=\min_{h'\in\text{deployed}} d(h_j,h')$ from Eq.~\eqref{eq:dist}. Records dominated on (utility, cost) are dropped first, so the archive acts as stepping stones~\cite{dgm}. The proposer sees only retrieved records and may apply one \emph{auditable typed edit} per candidate, drawn from a closed set (workers, edges, one role field, budget, aggregation, verification; Table~\ref{tab:space}). Each proposal is a machine-readable diff with a hypothesis, mechanism, and anticipated failure mode (Figure~\ref{fig:overview}, \ding{183}). We exclude unrestricted code generation: free-form self-rewriting~\cite{dgm,alphaevolve} would change the candidate distribution and evaluation surface at once, confounding calibration attribution.\looseness=-1{}

\paragraph{Paired evaluation and interventional credit assignment.}
Candidate and incumbent run on the \emph{same} packets, seeds, market paths, and tool responses; pairing removes path luck and sharply reduces the variance of the improvement estimate. For component $i$ (worker, edge, verification node, budget increment) we estimate Shapley-style interventional credit
\begin{equation}
\phi_i(h) \;=\; \mathbb{E}_{\pi}\Big[\, J\big(\mathrm{Pre}_\pi(i) \cup \{i\}\big) - J\big(\mathrm{Pre}_\pi(i)\big) \,\Big],
\label{eq:credit}
\end{equation}
where $\pi$ is a uniform random permutation of $C(h)$ and $\mathrm{Pre}_\pi(i)$ the components preceding $i$, estimated over a fixed permutation budget (Table~\ref{tab:hparams}) under common random numbers, counted in search compute. The removal operator is \emph{compiler-valid}: dropping $i$ re-routes orphaned edges to the aggregator, substitutes a neutral pass-through for a removed verifier, and holds the token budget fixed. Infeasible coalitions take the incumbent's projected utility, so $J(S)$ is always defined. Because the imputation is a design choice shrinking the credit of components whose removal breaks compilation, these are \emph{compiler-defined Shapley-style credits} rather than canonical Shapley values (alternative rule: Appendix~\ref{app:sim}). We do not report the Monte Carlo standard error of $\hat\phi$, so the evaluated outcome is the \emph{rank} correlation; individual magnitudes should not be read quantitatively. Credits from round $k$ enter the archive only at round $k{+}1$, never within the round that generated them. The pool $\mathcal{H}_k$ is therefore fixed before the shadow evidence that evaluates it, the pre-generation condition of Proposition~\ref{prop:fp}. Attribution is \emph{interventional}, so a persuasive worker that adds no counterfactual utility receives no credit. Appendix~\ref{app:related} positions this against observational textual backpropagation~\cite{trace,gepa}.\looseness=-1{}

\paragraph{Adversarial and negative-control evaluation.}
The evaluator applies five intervention families (dropout, corrupted workers, stale data, cost shocks, descriptor corruption; Appendix~\ref{app:sim}). Two negative controls are central: under \textbf{clone-collapse} all workers share identical information and competence, so a sound meta-agent should contract~\cite{illusion}; permuting $z_k$ (\textbf{shuffled-regime}) should make any conditional advantage vanish (Table~\ref{tab:ablate}).\looseness=-2{}


\paragraph{Conservative promotion with calibrated simultaneous correction.}
For candidate $c$, let $D_{c,b}=\Jc_{k,b}(c;h_k)-\Jc_{k,b}(h_k;h_k)$ be the paired improvement of Eq.~\eqref{eq:certified} on effective block $b$, where $B_{\mathrm{eff}}$ counts approximately independent blocks (circular block bootstrap~\cite{kunsch}, pre-selected length). At update $k$ the archive and pool are frozen at cutoff $t_k$; the pre-designated window $W_k$ is scored without revealing outcomes to the proposer, and labels and credits enter $\mathcal D_{k+1}$ only afterwards. Successive 26-week windows overlap at monthly cadence, so later rounds may reuse earlier observations, but no round adapts its pool to its own held-out $W_k$ (Appendix~\ref{app:repro}). The \emph{implemented} gate applies an empirically calibrated, variance-penalized simultaneous score,
\begin{equation}
\Score_k(c) \;=\; \bar D_c \;-\; r_{k,c},
\qquad
r_{k,c} \;=\; \sqrt{\frac{2\,\widehat V_{k,c}\,\log\!\big(2|\mathcal{H}_k|/\delta_k\big)}{B_{\mathrm{eff}}}},
\label{eq:score}
\end{equation}
treating the tightly concentrated paired deltas (empirical range ${\approx}0.04$ under common random numbers) as sub-Gaussian; the idealized Bernstein analysis appears in Appendix~\ref{app:proof}.
Here $\widehat V_{k,c}$ is the sample variance of the $D_{c,b}$. Because both $\widehat V_{k,c}$ and $B_{\mathrm{eff}}$ are estimated, $\Score_k$ is \emph{not a confidence bound}; it is calibrated empirically (Figure~\ref{fig:gate}; Remark~1, Appendix~\ref{app:proof}). Among feasible candidates, the gate promotes the highest-scoring one (ties broken toward fewer workers) only if $\Score_k(c)>\epsilon$; otherwise it retains the incumbent. Promotion first enters a simulated, limited-authority canary stage alongside the incumbent, with pre-specified rollback triggers.

\noindent The result below assumes a \emph{valid} confidence radius, which Eq.~\eqref{eq:score} does not provide. We state it because it forces the design conditions.

\begin{proposition}[Idealized conditional family-wise false-promotion result; \emph{not} a property of Eq.~\eqref{eq:score}]
\label{prop:fp}
Fix the candidate pool $\mathcal{H}_k$ before the shadow sample is drawn ($\mathcal{H}_k$ measurable w.r.t.\ information prior to $W_k$), and let $r^{\mathrm{valid}}_{k,c}$ be any valid $1-\delta_k/|\mathcal H_k|$ confidence radius for candidate $c$'s mean block improvement. Define the \emph{idealized} lower confidence bound $\mathrm{LCB}^{\mathrm{ideal}}_k(c)=\bar D_c-r^{\mathrm{valid}}_{k,c}$. Then, with probability at least $1-\delta_k$, every candidate promoted under $\mathrm{LCB}^{\mathrm{ideal}}_k(c)>\epsilon$ has expected \emph{replay} improvement exceeding $\epsilon$; choosing $\delta_k=\delta/K$ (or $\delta_k=6\delta/\pi^2k^2$) bounds the probability of any false promotion across updates by $\delta$.
\end{proposition}
\noindent\emph{Departures.} The proof is a standard simultaneous-concentration argument~\cite{hcpi,spibb,consbandit}. Eq.~\eqref{eq:score} breaks three of its premises (Appendix~\ref{app:proof}): overlapping windows break cross-update disjointness; the variance-only penalty drops the Bernstein range correction and plugs in an \emph{estimated} block count, so no finite-sample theorem applies; and the deltas are winsorized. We therefore call the gate \emph{empirically calibrated} (Figure~\ref{fig:gate}). A \emph{sealed-protocol} rerun removes cross-update overlap and realizes 4.8\% at the 5\% target. It does not validate Proposition~\ref{prop:fp}, which concerns \emph{replay} improvement and says nothing about a misspecified objective (RQ7). A real deployment would inflate the margin to $\epsilon+\xi_k$; every number here uses $\xi_k=0$, and nothing was run live.\looseness=-1{}

\textbf{Scoping.} \emph{Within} an update, $\delta$ is split across the $|\mathcal H_k|$ candidates and budgeted at $\delta=0.05$ \emph{per update}, which is the denominator that the 3.4\% rate and Figure~\ref{fig:gate} report against. Sequence-level control would require $\delta_k=\delta/K$ or the anytime schedule, neither activated for Track~A. The implied per-stream bound is \emph{vacuous}, so RQ4's per-stream rate is covered by no 5\% guarantee (Appendix~\ref{app:proof}).\looseness=-1{}


\section{\bench: benchmarking meta-level adaptation}
\label{sec:bench}

No single backtest can establish meta-agent quality, so \bench{} triangulates controlled capability (Track~A), external validity (Track~B), and adversarial governance (Track~C) over a shared worker library, risk layer, and budget accounting (Figure~\ref{fig:bench}).

\textbf{Track~A: controlled regime environment.} The primary controlled benchmark is a hidden-state factor process, a Markov regime-switching model after~\cite{hamilton}. The simulator specifies regime-dependent worker competence, so a surrogate-optimal architecture is computable. The regime evolves, a signal about next-period factors is emitted \emph{before} the decision, and the return realizes afterward:
\begin{equation}
\begin{aligned}
&s_{t+1}\sim P(\cdot\mid s_t), \qquad q_t,\ \nu_{i,t},\ \eta_{i,t}\in\mathcal F_t, \qquad
\nu_{t+1},\ \epsilon_{t+1}\ \perp\ \mathcal F_t,\\
&u_{t+1}=A_{s_t}q_t+\nu_{t+1}, \qquad
v_{i,t}=C_{i,s_t}q_t+\nu_{i,t}, \qquad
f_{t+1}=\mu_{s_t}+u_{t+1},\\
&y_{i,t}=H_{i,s_t}\mu_{s_t}+G_{i,s_t}v_{i,t}+\eta_{i,t}, \qquad
r_{t+1}=B_{s_t}f_{t+1}+\epsilon_{t+1}.
\end{aligned}
\label{eq:sim}
\end{equation}
Here $q_t$ is the pre-decision predictable component; worker-observation noises $\nu_{i,t},\eta_{i,t}$ are $\mathcal F_t$-measurable at decision time, while the time-$(t{+}1)$ innovations $\nu_{t+1},\epsilon_{t+1}$ are independent of $\mathcal F_t$ (all sampled independently given the regime). $A_{s_t}$ and $C_{i,s_t}$ induce the worker--future-shock correlation without exposing a future-indexed quantity; $G_{i,s_t}$ sets signal strength; workers never see $f_{t+1}$, $r_{t+1}$, or oracle labels. The tensors $H,G$ define hidden regime-dependent competence (some specialists useful only in some regimes; one persuasive but anti-correlated). Knowing them, the evaluator computes a \emph{surrogate dynamic-oracle sequence} on the same schedule and budget, charging switching cost rather than taking the per-regime best. That makes regret, selection accuracy, and credit fidelity computable.\looseness=-1{}

\textbf{LLM workers in Track~A.} A worker is an LLM specialist: it receives a rendered \emph{evidence packet} built from its noisy signal $y_{i,t}$, the context $z_k$, and role instructions, and returns $o_{i,t}$. The simulator sets signal quality via $H,G$, while the LLM supplies calibration, disagreement, and failure modes (over-confidence, susceptibility to the ``persuasive'' anti-correlated worker), so Track~A tests \emph{orchestration} (worked packet in Appendix~\ref{app:sim}). The evaluator observes $s_t$ only to score the surrogate oracle; every method sees the same $z_k$ (no-leakage conditions: Eq.~\eqref{eq:sim}).\looseness=-1{}

\textbf{Track B: market replay; Track C: worker-fleet governance.} Track~B is a secondary, underpowered external-validity check on historical data, using date-controlled and counterfactual structural replay against documented LLM-backtest leakage~\cite{profitmirage,lookahead,memctrl} (Appendix~\ref{app:sim}). Track~C is a market-noise-free suite run under a budgeted meta-adversary, following the compromised-insider threat model of~\cite{mama}. It tests whether the meta-agent detects a corrupted worker, isolates the affected path, picks a minimal repair, avoids leaking restricted context, and rolls back after a simulated governance incident.\looseness=-1{}


\section{Experimental protocol}
\label{sec:protocol}

\textbf{Baselines.} Seven Track~A methods: a strong single agent with self-consistency under the same per-decision \emph{deployment} budget cap; a fixed full committee; a rule-based regime router; a LinUCB bandit~\cite{linucb} over a 15-architecture menu (partially logged; Table~\ref{tab:linucb}); AFlow-style workflow search~\cite{agentsquare,aflow}; MaAS/AutoMaAS-style adaptive selection~\cite{maas,automaas}, the strongest adaptive competitor and what ``best adaptive baseline'' denotes throughout; and GEPA-style prompt-only optimization~\cite{gepa}. All are scored against the same surrogate dynamic oracle and share data packets, $z_k$, worker library, risk layer, model pool, and equal \emph{search-inclusive budget caps} (Shapley and paired-replay compute counted), with no information advantage. These are our reimplementations, not official releases, so every superiority claim is scoped to \emph{common-budget reimplementations under a shared risk layer}. Each method's defining search mechanism is retained; only what the shared environment forces is changed (retained/changed detail in Appendix~\ref{app:sim}). Routing the best baseline's proposals through our gate still reaches 0.45 OOD utility against \cmas's 0.48 (Table~\ref{tab:ablate2}), bounding the risk that a weak reimplementation drives the result.\looseness=-1{}

\textbf{Research questions.}\looseness=-1{} \textbf{RQ1} regret and cost-adjusted utility; \textbf{RQ2} interventional worker competence vs.\ evaluator noise; \textbf{RQ3} transfer to unseen regimes, libraries, and backbones (Definition~\ref{def:metagen}); \textbf{RQ4} gate calibration under serial dependence; \textbf{RQ5} sparsity when extra agents add no value; \textbf{RQ6} recovery from compromised workers; \textbf{RQ7} drift under a misspecified (return-only) evaluator.

\textbf{Metrics.} Track~A: normalized mean surrogate dynamic-oracle regret; selection accuracy; \emph{adaptation delay} (updates until post-shift regret returns within $\epsilon_{\mathrm{delay}}$, distinct from the margin $\epsilon$); credit rank correlation. Track~B: CEQ, Sharpe, MDD, CVaR, turnover, factor $\alpha$. All tracks: cost, size, churn, violations, rollback precision/recall, and \emph{false-promotion rate}. That rate is the fraction of per-update \emph{decisions} promoting a candidate whose simulator-evaluated mean improvement is $\le\epsilon$ (undefined where a method never switches).\looseness=-1{}

\textbf{Statistics.} Nested walk-forward with an untouched final test; search is confined to training/validation intervals. The unit of replication is a \emph{stream} (96 $=$ 12 seeds $\times$ 8 schedules), with paired stream-level block-bootstrap CIs as the primary inference and schedule-clustered hierarchical intervals as a sensitivity analysis (Tables~\ref{tab:main}, \ref{tab:ablate2}--\ref{tab:ablate}). With only eight top-level schedules, clustered intervals are read cautiously and significance rests on the \emph{paired} stream-level difference under common random numbers; clustering widens intervals without overturning the ablations (e.g.\ no-paired-evaluation OOD utility $[0.42,0.48]$; Table~\ref{tab:ablate}). Paired \cmas{}-vs-best-baseline differences in regret and OOD utility exclude zero after Holm correction. The full pairwise matrix is omitted for space (seed separation in Appendix~\ref{app:repro}).\looseness=-1{}

This \emph{simulator-evaluated mean improvement} is not directly observed: the evaluator estimates it from 50 Monte Carlo paths per candidate (SE $<0.004$; ties within $\pm0.005$ of $\epsilon$ labeled false conservatively). Score form, block length, winsorization, $\epsilon$, and pool size were frozen at validation before the untouched final test (Tables~\ref{tab:timeline}--\ref{tab:splits}; configuration Table~\ref{tab:hparams}).\looseness=-1{}


\section{Results and diagnostic analyses}
\label{sec:results}

\begin{table}[t]
\centering
\caption{\textbf{Primary meta-level results} (Track~A; 96 streams; $\pm$95\% stream-level block-bootstrap CIs). Search rows are our common-budget ``-style'' reimplementations. Cost is \emph{deployment} inference only and is \emph{not} a net saving: \cmas{} is heaviest on search-inclusive tokens (Table~\ref{tab:cost}). ``n/a'' $=$ never updates, so false promotion is undefined, not zero.}
\label{tab:main}
\footnotesize
\setlength{\tabcolsep}{5.4pt}
\begin{tabular}{lccccc}
\toprule
\textbf{Method} & Regret $\downarrow$ & OOD utility $\uparrow$ & False promo. $\downarrow$ & Team size & Cost $\downarrow$ \\
\midrule
Single agent + SC        & 0.226$\pm$.022 & 0.28$\pm$.05 & n/a & 1.0 & 1.00 \\
Fixed committee          & 0.183$\pm$.017 & 0.32$\pm$.03 & n/a & 6.0 & 2.40 \\
Rule router              & 0.158$\pm$.016 & 0.38$\pm$.03 & 0.071 & 3.5 & 1.45 \\
AFlow-style              & 0.137$\pm$.014 & 0.41$\pm$.05 & 0.093 & 4.7 & 2.05 \\
MaAS/AutoMaAS-style      & 0.116$\pm$.013 & 0.44$\pm$.03 & 0.082 & 4.0 & 1.83 \\
GEPA prompt-only         & 0.166$\pm$.019 & 0.36$\pm$.03 & 0.063 & 4.0 & 1.55 \\
\midrule
\textbf{\cmas{} (ours)}  & \textbf{0.104$\pm$.011} & \textbf{0.48$\pm$.05} & \textbf{0.034} & 3.4 & 1.72 \\
\bottomrule
\end{tabular}
\end{table}

\begin{figure}[t]
\centering
\begin{subfigure}[b]{0.40\textwidth}
  \includegraphics[width=\textwidth]{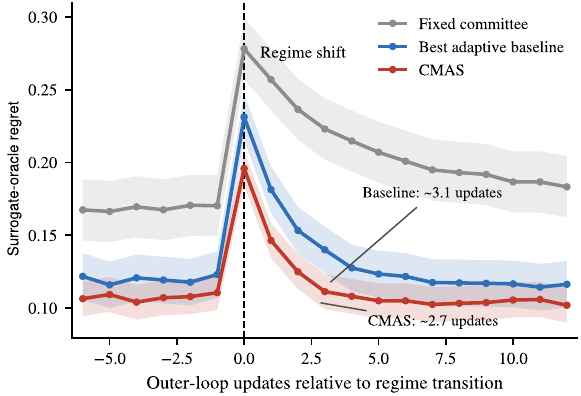}
\end{subfigure}\hfill
\begin{subfigure}[b]{0.40\textwidth}
  \includegraphics[width=\textwidth]{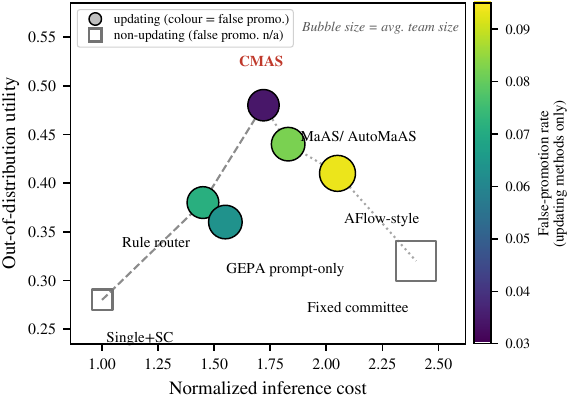}
\end{subfigure}
\caption{\textbf{Headline diagnostics.} \emph{Left:} regret around evaluator-known regime transitions, 95\% bands; \cmas{} recovers in $\sim$2.7 steps. \emph{Right:} utility--cost frontier on \emph{deployment-only} cost (\cmas{} is heaviest on search-inclusive tokens; Table~\ref{tab:cost}). Bubble size is average team size. Color shows false-promotion rate for updating methods; static methods make no promotion decisions, so they are drawn as hollow grey squares outside the color scale.\looseness=-1{}}
\label{fig:headline}
\end{figure}

\textbf{RQ1: Adaptation and cost-adjusted utility.} Among adaptive methods, \cmas{} attains the lowest regret, adaptation delay, and false promotion, and the highest OOD utility and credit fidelity (Tables~\ref{tab:main},~\ref{tab:secondary}). Versus the MaAS/AutoMaAS-style reimplementation, regret fell $0.116\to0.104$ (10.3\%), OOD utility rose $0.44\to0.48$, credit correlation $0.32\to0.41$, deployment inference cost was 6.0\% lower, and recovery took 2.7 updates vs.\ 3.1 (Figure~\ref{fig:headline}). Paired differences exclude zero after Holm correction: regret $-0.012$ ($[-0.020,-0.004]$), OOD utility $+0.038$ ($[+0.010,+0.066]$). \cmas{} is nevertheless heaviest in search-inclusive tokens (Table~\ref{tab:cost}); the frontier claim is deployment-only. Track~B is directionally consistent but underpowered ($N{=}18$; Table~\ref{tab:replay}); the LinUCB menu also trails (Table~\ref{tab:linucb}).\looseness=-1{}

\textbf{RQ2: Interventional fidelity.} Exact surrogate-oracle selection is 63.8\% vs.\ 59.1\%, credit correlation $\rho=0.41$ vs.\ 0.32 (per-stream median 0.41, CI $[0.35,0.47]$); selection rotates with regime, and counterfactually useless persuasive workers get low credit (Figure~\ref{fig:dynamics}).

\textbf{RQ3: Transfer.} Utility retention is 83.5\% on unseen regime mixtures (78.5\% baseline) and 81.2\% with two new specialists after four updates; backbone-switch loss is 7.8\% vs.\ AFlow-style's 11.8\% (Table~\ref{tab:transfer}). Held-out false promotion rises 3.4\%$\to$4.1\%. Strict safety generalization is not established. Shuffled descriptors remove the advantage.\looseness=-1{}

\textbf{RQ4: Promotion calibration.} False promotions are 3.4\% of 2,112 updates and 5.86\% of actual promotions; the incumbent is retained on 42\%. These are \emph{near-neutral} updates: 1.2\% of decisions are full-$J$ reductions and 0.4\% are negative-$\Jc$ promotions (Table~\ref{tab:defs}); 11.4\% of streams contain at least one false promotion. The sealed rerun yields 4.8\%; removing simultaneous correction or pairing raises the rate to 8.4\% or 6.1\% (Table~\ref{tab:ablate}). No full-utility, canary-harm, or sequence-level guarantee is claimed.\looseness=-1{}

\begin{figure}[t]
\centering
\includegraphics[width=0.58\textwidth]{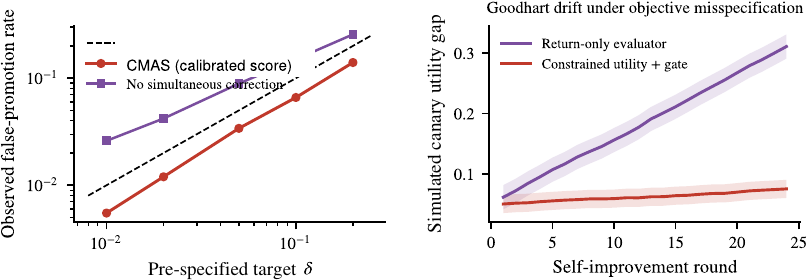}
\caption{\textbf{Promotion calibration and Goodhart drift.} Held-out final test (2,112 decisions, 96 streams); all plotted parameters frozen at validation (Table~\ref{tab:splits}), so neither panel is post-hoc fitted. \emph{Left:} realized false promotion vs.\ pre-specified $\delta$, observed below the nominal target on the tested distributions. This is empirical calibration; the score is not a valid confidence bound (Remark~1). \emph{Right:} one objective-misspecification probe illustrating a failure mode.}
\label{fig:gate}
\end{figure}

\textbf{RQ5 (sparsity).} Under clone-collapse \cmas{} contracts 3.4$\to$1.6 (4.6 without parsimony), a negative control rather than a general guarantee. \textbf{RQ6 (governance).} \emph{Simulated} recovery from a budgeted adversary is 78\%, rollback precision/recall 0.85/0.80 (Table~\ref{tab:gov}). \textbf{RQ7 (Goodhart).} A return-only objective develops a growing gap between simulated canary and deployed-configuration utility, unlike the constrained evaluator (Figure~\ref{fig:gate}); one misspecification is tested. Holding evaluator and gate fixed, proposal quality raises OOD utility 0.38$\to$0.48 while false promotion stays 3.2--3.5\% (Table~\ref{tab:ablate2}), separating search quality from calibration.\looseness=-1{}


\section{Discussion, limitations, and responsible use}
\label{sec:discussion}

\textbf{Evidence and limitations}, ordered by how much each bounds the conclusions. \emph{Statistical:} the implemented score falls outside Proposition~\ref{prop:fp} and certifies only $\Jc$. \emph{Benchmark:} the simulator is author-designed, so Track~A shows internal validity only, and Track~B is underpowered. \emph{Baseline:} comparisons are common-budget ``-style'' reimplementations. \emph{Scope:} one domain, an assumed informative $z_k$, and exact-architecture matching that can mark a near-optimal graph wrong, so regret is primary. \emph{Deployment:} the canary stage and rollback are simulated; nothing was run live. Claims rest on surrogate-oracle regret, negative controls, calibrated promotion, and held-out transfer, not backtested return~\cite{profitmirage,lookahead}.\looseness=-1{}

\paragraph{Responsible-use statement.} (Extended version in Appendix~\ref{app:responsible}.)
\emph{Intended use:} offline or shadow evaluation of \emph{bounded} architecture changes, as decision support for authorized operators, not autonomous authorization of high-impact actions, and over a closed search space (Table~\ref{tab:space}).
\emph{Stakeholders and harms:} updates affect operators, users of the managed system, owners of the data and tools its workers touch, downstream decision recipients, and here investors. A wrong selection can install an overconfident worker, a correlated team, an evaluator-tuned architecture, or a verifier that blocks useful actions or certifies unsafe ones.
\emph{Controls and authority:} risk projection, authorization, accounting, and data cutoffs sit outside the LLM boundary (Figure~\ref{fig:overview}); hard checks reject ineligible candidates before the gate; promotion is reversible. The owner sets objective, $\epsilon$, limits, and triggers, while sign-off, halt, and audit ownership should rest with a reviewer independent of the proposer's authors. We implement the technical controls and specify the human ones. Those we did not evaluate.
\emph{Misuse and residual risk:} the loop is objective-agnostic and could optimize fleets for surveillance or manipulation, or search around a misspecified evaluator. A candidate can pass the gate and still reduce full $J$ (1.2\% here); sequence-level error is uncontrolled; simulated canary-stage harm preceded rollback in one stream (Appendix~\ref{app:fail}); sign-off can decay into rubber-stamping.\looseness=-1{}


\section{Conclusion}
\label{sec:conclusion}
\cmas{} treats meta-agent search as reversible sequential deployment, and \bench{} makes ``did it adapt for the right reason?'' measurable. Together they give a bounded, auditable framework for deciding when to replace an incumbent. That is not proof that meta-agent updates are safe.\looseness=-1{}

\appendix

\section{Proof of Proposition~\ref{prop:fp} (false-promotion control)}
\label{app:proof}

\begin{table}[tbp]
\centering\small
\caption{\textbf{Terminology used for promotion outcomes.}}
\label{tab:defs}
\begin{tabular}{lp{9.4cm}}
\toprule
Term & Meaning \\
\midrule
Certified statistic $\Jc$ & additive block statistic in Eq.~\eqref{eq:certified}; excludes variance/CVaR/MDD \\
Full utility $J$ & complete scalarization incl.\ CVaR/MDD penalties; risk terms act as eligibility filters \\
Promotion & candidate replaces incumbent after $\Score_k(c)>\epsilon$ and eligibility pass \\
Canary stage & simulated post-promotion, limited-authority run alongside the incumbent with pre-specified rollback triggers \\
Rollback & reversion to the incumbent when a trigger fires \\
False promotion & promoted candidate whose simulator-evaluated mean improvement in $\Jc$ is $\le\epsilon$; near-neutral, and not necessarily a negative-$\Jc$ promotion \\
Negative-$\Jc$ promotion & promoted candidate whose simulator-evaluated mean improvement in $\Jc$ is below zero (0.4\% of all decisions) \\
Full-$J$ reduction & promotion improving $\Jc$ but reducing full $J$ within eligibility (1.2\% of decisions) \\
Simulated canary-stage harm & adverse outcome observed during the simulated canary run, before a rollback trigger fires; distinct from all three statistics above and not measured by the gate \\
Per-update rate & false promotions / all update decisions (3.4\% overlapping protocol; 4.8\% sealed) \\
Conditional rate & false promotions / actual promotions (5.86\%) \\
Per-stream family-wise rate & streams with $\ge$1 false promotion / all streams (11.4\%) \\
\bottomrule
\end{tabular}
\end{table}

For each candidate $c \in \mathcal{H}_k$, let $\Delta_c = \mathbb{E}[D_{c,b} \mid \mathcal{H}_k]$. By the assumed concentration bound,
\[
\Pr\!\big(\Delta_c < \bar D_c - r_{k,c} \,\big|\, \mathcal{H}_k\big) \;\le\; \frac{\delta_k}{|\mathcal{H}_k|},
\]
where, for this idealized statement, $r_{k,c}$ is the empirical-Bernstein radius (variance term of Eq.~\eqref{eq:score} plus the range correction $3M\log(2|\mathcal H_k|/\delta_k)/B_{\mathrm{eff}}$), valid for improvements bounded in $[-M,M]$ with sample variance $\widehat V_{k,c}$ over $B_{\mathrm{eff}}$ approximately independent blocks~\cite{anytime}. The \emph{implemented} score is the variance-only form of Eq.~\eqref{eq:score}, which is not itself a valid radius (Remark~1). Writing $\mathrm{LCB}^{\mathrm{ideal}}_k(c)=\bar D_c-r^{\mathrm{valid}}_{k,c}$, a union bound over the finite candidate pool implies that, with probability at least $1-\delta_k$, $\Delta_c\ge\mathrm{LCB}^{\mathrm{ideal}}_k(c)$ simultaneously for all candidates; hence any candidate promoted under $\mathrm{LCB}^{\mathrm{ideal}}_k(c)>\epsilon$ satisfies $\Delta_c>\epsilon$. A second union bound over $K$ updates with $\delta_k = \delta/K$ yields the global statement.

For serially dependent replay data, $B_{\mathrm{eff}}$ must come from a valid block construction. If the mixing or block assumptions fail, the guarantee does not apply. This condition is stress-tested empirically in Figure~\ref{fig:gate} (left); a heavy-tail/autocorrelation calibration suite is left to future work.

The configuration constant $M{=}0.15$ bounds the \emph{raw, unpaired} utility scale; under common random numbers candidate and incumbent share the same shocks, so the paired deltas $D_{c,b}$ concentrate far more tightly, with empirical range ${\approx}0.04$. Empirically, the median $B_{\mathrm{eff}}$ is $5.8$ and $M=0.15$, and implemented penalties fall in $r_{k,c}\in[0.014,0.032]$. Promoted candidates have $\bar D_c\in[0.028,0.048]$, while rejected candidates peak near $\bar D_c\approx0.015$.

\emph{Implemented score versus idealized radius.} At these constants the Bernstein range correction alone evaluates to ${\approx}0.45$ and is therefore not the implemented quantity. Worked implemented-score example: $\widehat V_{k,c}=1.8\times10^{-4}$, $B_{\mathrm{eff}}=5.8$, $\delta_k=0.05$, $|\mathcal H_k|=8$ ($\log 320\approx5.768$) give $r_{k,c}\approx0.0189$; with $\bar D_c=0.045$, $\Score_k(c)=0.0261>\epsilon=0.02$, so the candidate is promoted. \qed

\medskip
\noindent\textbf{Remark 1 (idealized sub-Gaussian radius).} If the block improvements were $\sigma_c$-sub-Gaussian with \emph{known} proxy $\sigma_c$, the radius would reduce to $\sigma_c\sqrt{2\log(2|\mathcal H_k|/\delta_k)/B_{\mathrm{eff}}}$. A plug-in variance in that form is \emph{not} a valid radius; the empirical-Bernstein (or a self-normalized) bound is what licenses estimated variance. The implemented $\Score_k$ of Eq.~\eqref{eq:score} is exactly such a plug-in construction, which is why it is treated as empirically calibrated (Section~\ref{sec:method}).

\medskip
\noindent\textbf{Anytime-valid variant.} Replacing $\delta_k=\delta/K$ by $\delta_k = 6\delta/\pi^2 k^2$ gives $\sum_{k\ge 1}\delta_k=\delta$, so the family-wise guarantee holds for an \emph{unbounded} sequence of updates without pre-committing to $K$; a time-uniform confidence sequence~\cite{anytime} yields the same conclusion under martingale dependence. The sequence of pools $\mathcal H_1,\dots$ is itself adaptive (the proposer sees cross-round archive feedback). Formally, let $\mathcal F_{k-1} \supseteq \sigma(\mathcal H_1,\dots,\mathcal H_k,\,W_1,\dots,W_{k-1})$ be the information available before shadow window $W_k$ is drawn; because each $\mathcal H_k$ is $\mathcal F_{k-1}$-measurable, each round's radius is valid \emph{conditionally on} $\mathcal F_{k-1}$, and the union over rounds applies to the resulting conditional bounds. As stated in Section~\ref{sec:method} (Scoping), the reported Track~A runs budget $\delta$ per update and do not activate this schedule.

\medskip
\noindent\textbf{Idealized versus implemented.} Three premises of Proposition~\ref{prop:fp} are not established for the overlapping evaluation protocol: (i) cross-update disjointness fails because 26-week windows at monthly cadence overlap, so proposals condition on outcomes contained in earlier evaluation windows; (ii) the implemented score drops the Bernstein range correction and plugs an estimated $\widehat V_{k,c}$ and effective block count $B_{\mathrm{eff}}$ into a variance-only form, which carries no finite-sample concentration theorem under dependence (Remark~1); and (iii) the paired deltas are winsorized at their validation-calibrated 99\% quantile (binding in 1.2\% of blocks, mainly volatility shocks), while the configuration constant $M{=}0.15$ is a raw-utility-scale bound, loose by roughly $4\times$ on the paired-delta scale.

Gate sensitivity: block length 2/4/8 weeks gives realized FP 3.8/3.4/2.9\%; replacing the circular-block $B_{\mathrm{eff}}$ estimator with autocorrelation-based ESS gives 3.5\%. Removing the winsorization inflates $\widehat V_{k,c}$, widening the radius, lowering the promotion rate by ${\sim}4\%$, and raising realized false promotion to 4.1\%, still under target. The reported false-promotion rates under the overlapping evaluation protocol are empirical calibration.

The direct test of the idealized guarantee is a sealed-sample calibration study with fresh, non-overlapping evaluation paths per update. It realizes 4.8\% false promotion at the nominal 5\% target (vs.\ 3.4\% under the overlapping protocol), which shows that calibration is not an artifact of cross-window overlap. Because the implemented score was used, this is not a direct finite-sample validation of Proposition~\ref{prop:fp}. Ground-truth labels use 50 independent Monte Carlo paths per candidate (standard error $<0.004$ normalized units); candidates within $\pm0.005$ of $\epsilon$ are conservatively classified as false promotions; propagated label uncertainty contributes $<0.2$\,pp to the 3.4\% rate, and the sealed evaluation paths are disjoint from $W_k$ and independent across updates. Under that reported per-update budget, the implied worst-case per-stream bound over $K\approx22$ updates is vacuous ($K\delta>1$; ${\approx}0.68$ even under cross-update independence). The observed 11.4\% per-stream family-wise rate (RQ4) is far below what the configuration nominally permits, and above the per-update 5\% target, which applies to a different denominator.

\section{Pre-specified search space}
\label{app:space}
Table~\ref{tab:space} fixes the search space; every candidate edit selects values from these dimensions, and nothing outside them was searched.
\begin{table}[tbp]
\centering\small
\caption{\textbf{Pre-specified search space.}}
\label{tab:space}
\begin{tabular}{p{2.4cm}p{10.6cm}}
\toprule
\textbf{Dimension} & \textbf{Pre-specified values} \\
\midrule
Workers & any subset of six roles; per-role backbone is a separate model-assignment field (two open-weight options; twelve role--backbone instances) \\
Graph & DAG; $\le 10$ edges; no self-loops; permissions constrain edge types \\
Aggregation & parallel merge, confidence-weighted, sequential, debate, critic-then-aggregate \\
Models & one small open model, one stronger open model, one hosted reasoning model \\
Budget & $1\times$, $2\times$, $4\times$ token/tool budgets with a hard per-episode ceiling \\
Prompt mutation & one typed field per candidate; no hidden instruction changes \\
Verification & none, evidence validator, numerical recomputation, counterargument, dual-model critic \\
Promotion & calibrated simultaneous score, hard policy checks, parsimony tie-break, limited-authority canary stage, rollback snapshot \\
\bottomrule
\end{tabular}
\end{table}

\textbf{Chance baselines for selection accuracy} (for the RQ2 figures of Section~\ref{sec:results}). Uniform guessing over the surrogate oracle's realized 6-architecture support gives 16.7\%, the majority-architecture baseline 18.2\% (full-space chance far lower, Appendix~\ref{app:space}).

\section{Simulator specification and required checks}
\label{app:sim}
\textbf{Primary suite.} \emph{Five recurring macro-regime types} (Growth, Inflation, Stress, Recovery, Range; Figure~\ref{fig:motivation}) are arranged into \emph{eight distinct latent-regime schedules} (transition sequences), so the unit of replication is a \emph{stream} and 96 streams $=$ 12 simulation seeds $\times$ 8 schedules. \emph{Six worker roles} (trend, macro, value/carry, sentiment, liquidity/tail-risk, skeptic), each instantiated on two open-weight backbones drawn from the three-model pool (Appendix~\ref{app:space}), so the default library holds twelve specialists; the third, hosted model is reserved for the backbone-transfer test (RQ3). Each stream spans ${\sim}122$ weeks, including a 26-week \emph{unscored burn-in} that fills the first shadow window; the remaining 96 scored weeks yield $K\approx22$ monthly gate decisions, and streams include recurrent, abrupt, and gradual transitions. The Track~C governance suite injects six incident scenarios per stream: the five single-attack families of Section~\ref{sec:bench} and Table~\ref{tab:gov} (instruction corruption, anti-correlated worker, stale/missing data, inference-cost shock, worker dropout), plus one multi-attack combination sampled per stream. That gives $96\times 6 = 576$ governance episodes. The 5/10/25\,bps cost levels apply to Track~B replay, not to governance. The competence tensor includes specialists useful only jointly, redundant specialists, a cheap weak agent, an expensive strong agent, and a persuasive anti-correlated agent.

Oracle utility is computed by exhaustive search for small graphs and mixed-integer optimization for the larger space. The oracle is a \emph{surrogate}: it optimizes over the enumerated feasible subset of the pre-specified space (valid DAGs within budget) using the simulator's specified competence model. In that model, worker outputs enter through their analytic expected utility under the regime's factor distribution, marginalizing the simulator's noise terms; no actual LLM executions are involved. Prompt fields and verification nodes are fixed categorical choices in the MIP. Its optimality is exact for this surrogate model, not for arbitrary LLM behavior.

\textbf{Track B detail.} Historical evaluation uses two safeguards against documented LLM-backtest leakage~\cite{profitmirage,lookahead,memctrl}: \emph{date-controlled replay} restricted to information released after each model's documented training cutoff (for Mixtral the cutoff is undisclosed and January 2024 is used, so genuine post-training-cutoff status cannot be guaranteed for that model), and \emph{counterfactual structural replay} that fits a factor model, scrambles identities and calendars, and resamples factor/residual paths preserving cross-asset dependence. Agents never receive real tickers, dates, future documents, or web access. Performance is decomposed into market, style, and residual components so passive beta is not counted as skill, and publication lags and vintage data are frozen. The universe spans liquid equity, duration, credit, commodity, gold, real-estate, and international ETFs; decisions weekly, updates monthly (a quarterly variant is not evaluated here), costs 5/10/25\,bps. Transfer is evaluated on a held-out universe and a different model family. Track~B CEQ and factor $\alpha$ are reported in annualized percent throughout (e.g.\ \cmas{} CEQ $=7.8\%$); the cross-window standard deviation over the 18 replay windows is $\pm1.1$\,pp.

\textbf{Pre-specified analysis protocol.} We separate analyses \emph{reported in this paper} from the benchmark's standing evaluation suite. Reported here: strong single agent under the same per-decision deployment budget cap (Table~\ref{tab:main}); the random/evolutionary proposer inside the \cmas{} evaluator+gate (Table~\ref{tab:ablate2}); no/shuffled regime context and the component ablations (Table~\ref{tab:ablate}); identical-worker clones and different evaluator/worker family at test time; objective misspecification (return-only evaluator, RQ7); and search-inclusive cost (below). Not evaluated here, and therefore not claimed: Bayesian/change-point proposers, an oracle-context upper bound, candidate-pool scaling curves, and heavy-tail/autocorrelation gate calibration.

\textbf{Baselines: full Track~A roster} (summarized in Section~\ref{sec:protocol}). We compare against seven methods in Track~A: a strong single agent with self-consistency, evaluated under the same per-decision \emph{deployment} budget cap (no search budget; Table~\ref{tab:cost}); a fixed full committee; a rule-based regime router (rules in Appendix~\ref{app:sim}); a LinUCB bandit~\cite{linucb} on a fixed 15-architecture menu over $z_k$; AFlow-style workflow search~\cite{agentsquare,aflow}; MaAS/AutoMaAS-style adaptive selection~\cite{maas,automaas}, which is the strongest adaptive competitor and is what ``best adaptive baseline'' denotes in every figure and table; and GEPA-style prompt-only optimization~\cite{gepa} on the fixed manager--worker graph. Track~B adds risk parity. All Track~A methods are scored against the same surrogate dynamic oracle and share data packets, the context $z_k$, worker library, risk layer, model pool, and equal \emph{search-inclusive budget caps} (Shapley and paired-replay compute counted), with no information advantage. Realized usage differs under the cap: \cmas{} is highest in tokens though not wall-clock (Table~\ref{tab:cost}, Appendix~\ref{app:sim}). Because the environment mandates a shared risk layer and a common upper budget cap, published systems are reimplemented as ``-style'' variants (Appendix~\ref{app:sim}); Table~\ref{tab:ablate2} bounds the resulting risk for the gate comparison, Section~\ref{sec:discussion} for the search-quality comparison. Not run here: equal-weight/trend strategies, a human-designed manager--worker graph, and GPTSwarm/G-Designer-style topology search~\cite{gptswarm,gdesigner}; random and evolutionary proposers \emph{are} run inside the \cmas{} evaluator+gate (Table~\ref{tab:ablate2}).

\textbf{Baseline fidelity: retained vs.\ changed.} Every comparison method is a
common-interface reimplementation, not an official release, and all comparative claims
in the paper are scoped accordingly. \emph{Retained} for each method is its defining
search mechanism: AFlow-style keeps graph-edit search with lookahead; MaAS/AutoMaAS-style
keeps supernet optimization over discrete architectural choices; GEPA-style keeps
reflective prompt mutation with tournament selection (operators, depths and population
sizes below). \emph{Changed}, and only where the shared environment forces it: (i) the
identical 0.85M-token/update search-inclusive budget cap; (ii) the common typed worker
interface $o_{i,t}$, so that every method emits the same schema; (iii) the immutable risk
projection $\Pi_{\mathcal C_t}$ and authorization layer, which no method may bypass;
(iv) the common three-model pool and worker library; and (v) the identical per-decision
deployment budget cap. \emph{Information available} is the same for all methods: the
pre-decision descriptor $z_k$, the same data packets, seeds, market paths and tool
responses, and no access to the latent regime $s_t$ or to evaluator oracle labels. Two
measurements bound the residual risk that a weak reimplementation drives the headline
result: routing the strongest baseline's proposals through the \cmas{} evaluator and gate
still reaches 0.45 OOD utility against \cmas's 0.48 (Table~\ref{tab:ablate2}), and the
random/evolutionary proposer control in the same table separates proposal quality from
gate calibration.

\textbf{Baseline specifications.} The rule-based regime router maps $z_k$ to a hand-specified team via three conditional branches: \emph{Calm} (volatility descriptor $\le 0.15$): Trend $+$ Sentiment (2 specialists); \emph{Volatile} ($>0.15$): Tail-Risk $+$ Skeptic $+$ Macro $+$ Value/carry (4); \emph{Crisis} ($>0.30$ or drawdown trigger): those four plus Liquidity (5). Weighted by realized regime frequencies across the 8 schedules, the expected team size is 3.48, the 3.5 of Table~\ref{tab:main}. For the search baselines: AFlow-style runs DFS/BFS over 12 graph-edit operators with lookahead depth 3; MaAS/AutoMaAS-style optimizes a supernet with 15 discrete architectural choices via REINFORCE with a moving-average baseline; GEPA-style runs reflective prompt mutation with population 20 and tournament selection. All are capped at the identical 0.85M-token/update search-inclusive budget as \cmas{}.

The infeasible-coalition imputation used in credit estimation shrinks the estimated credit of components whose removal breaks compilation toward the incumbent's level rather than measuring their true marginal value. The direction and size of the resulting bias depend on the repair rule, which we treat as a design choice (an alternative repair rule is evaluated in Appendix~\ref{app:sim}). \textbf{Alternative credit repair.} A drop-and-penalize rule (infeasible coalitions scored 0) raises credit variance 14\% without materially changing selection accuracy (63.1\% vs.\ 63.8\%). \textbf{Proposer-component ablations} (OOD utility; full proposer 0.48): removing archive retrieval drops it to 0.41, removing credit feedback to 0.44, and removing regime conditioning to 0.44, so retrieval carries the largest share of the proposer's edge.

\begin{table}[tbp]
\centering\small
\caption{\textbf{Closest search and adaptation methods.} Entries summarize first-class mechanisms in the cited descriptions.}
\label{tab:compare-search}
\begin{tabular}{L{2.6cm}L{3.2cm}L{2.3cm}L{4.2cm}}
\toprule
\textbf{Method} & \textbf{Search object / proposal} & \textbf{Adaptation scope} & \textbf{Primary distinction from \cmas{}} \\
\midrule
ADAS / AFlow / AgentSquare~\cite{adas,aflow,agentsquare} & executable programs or workflows & offline/static or per task & no repeated paired incumbent replacement \\
MaAS / AutoMaAS~\cite{maas,automaas} & architecture and resources & query-level & query routing rather than deployment-time promotion calibration \\
GEPA / Darwin G\"odel Machine~\cite{gepa,dgm} & prompts or executable code & iterative/open-ended & population fitness rather than paired replay and rollback \\
Adaptive Auto-Harness~\cite{adaptiveharness} & harness tree and stateful evolver & open-ended non-stationary streams & solve-time routing/human steering; no explicit update-level false-promotion target \\
Continual Harness / AutoRAS~\cite{contharness,autoras} & online or robustness-aware harnesses & continual/robust & no explicit paired incumbent gate with measured false-promotion error \\
\textbf{\cmas{}} & typed architecture and archive-conditioned edits & sequential deployment & paired replay, calibrated score, switching cost, canary stage, rollback \\
\bottomrule
\end{tabular}
\end{table}

\begin{table}[tbp]
\centering\small
\caption{\textbf{Closest safety and deployment-control methods.} Related ingredients are acknowledged; \cmas{} contributes their deployment-level combination.}
\label{tab:compare-safety}
\begin{tabular}{L{2.8cm}L{4.0cm}L{5.5cm}}
\toprule
\textbf{Method} & \textbf{First-class safety mechanism} & \textbf{Relation to \cmas{}} \\
\midrule
Gated semantic QD~\cite{gatedqd} & deterministic candidate-quality gate & no paired incumbent replay, rollback, or update-level calibration \\
HARBOR~\cite{harbor} & posterior chance-constrained safety check & no repeated incumbent replacement or rollback \\
HarnessAudit~\cite{harnessaudit} & full-trajectory boundary and information-flow audit & complementary audit surface; does not optimize sequential promotion \\
SafeHarness~\cite{safeharness} & lifecycle verification, privilege control, adaptive degradation, rollback & complementary execution security; no false-promotion estimand \\
\textbf{\cmas{}} & paired calibrated promotion, hard eligibility, canary stage, rollback & focuses on whether and when to replace the incumbent under drift \\
\bottomrule
\end{tabular}
\end{table}

The LinUCB menu bandit runs LinUCB~\cite{linucb} on context $z_k$ over a fixed menu of 15 architectures, enumerated by taking the 5 most frequently Pareto-optimal architectures from historical (pre-test) runs and crossing them with the 3 pre-specified budget levels ($1\times,2\times,4\times$; Appendix~\ref{app:space}), with exploration parameter $\alpha=1.0$.
\begin{table}[tbp]
\centering\small
\caption{\textbf{Partially logged LinUCB menu baseline.} Only the two metrics below were retained; it is therefore excluded from the fully populated primary comparison in Table~\ref{tab:main}.}
\label{tab:linucb}
\begin{tabular}{lcc}
\toprule
Method & Normalized mean surrogate dynamic-oracle regret $\downarrow$ & OOD utility $\uparrow$ \\
\midrule
LinUCB menu bandit & 0.142 & 0.38 \\
\bottomrule
\end{tabular}
\end{table}

\begin{table}[tbp]
\centering\small
\caption{\textbf{Proposer-value and gate decomposition.} Proposer varied; evaluator, credit, and gate fixed. ``$+$ \cmas{} gate'' routes that baseline's proposals through our gate. Hierarchical (schedule-clustered) 95\% CIs on OOD utility, top to bottom: $[.46,.50]$, $[.43,.47]$, $[.40,.44]$, $[.36,.40]$. Proposer-component ablations: Appendix~\ref{app:sim}.\looseness=-1{}}
\label{tab:ablate2}
\footnotesize
\begin{tabular}{lccc}
\toprule
\textbf{Proposer / config.\ (fixed evaluator+gate)} & OOD util.\ $\uparrow$ & Dyn.\ regret $\downarrow$ & False promo.\ $\downarrow$ \\
\midrule
LLM proposer (full \cmas{}) & \textbf{0.48} & \textbf{0.104} & 0.034 \\
Best adaptive baseline $+$ \cmas{} gate & 0.45 & 0.113 & 0.035 \\
Evolutionary candidates     & 0.42 & 0.129 & 0.033 \\
Random candidates           & 0.38 & 0.151 & 0.032 \\
\bottomrule
\end{tabular}
\end{table}

\begin{table}[tbp]
\centering\small
\caption{\textbf{Secondary Track~A metrics}, held out of the headline table for readability. Adaptation delay is right-censored at the eight-update event-study horizon; exact architecture matching is used for credit/selection diagnostics, while regret remains the primary oracle-relative metric.}
\label{tab:secondary}
\begin{tabular}{lccc}
\toprule
\textbf{Method} & Adaptation delay $\downarrow$ & Credit $\rho$ $\uparrow$ & Notes \\
\midrule
Single agent + SC & 7.3$\pm$.8 & -- & static \\
Fixed committee & 6.2$\pm$.6 & 0.11$\pm$.06 & static \\
Rule router & 4.3$\pm$.5 & 0.18$\pm$.05 & hand-routed \\
AFlow-style & 3.5$\pm$.4 & 0.25$\pm$.05 & search \\
MaAS/AutoMaAS-style & 3.1$\pm$.3 & 0.32$\pm$.04 & strongest adaptive baseline \\
GEPA prompt-only & 4.6$\pm$.5 & 0.18$\pm$.05 & fixed graph \\
\textbf{\cmas{}} & \textbf{2.7$\pm$.3} & \textbf{0.41$\pm$.05} & paired calibrated promotion \\
\bottomrule
\end{tabular}
\end{table}

\textbf{Search-inclusive cost.} The ``Cost'' column in Table~\ref{tab:main} is \emph{deployment-only}; Table~\ref{tab:cost} reports realized search-inclusive usage per update. \cmas{} is the heaviest token consumer, while its wall-clock is below the sequential graph-search baselines because paired replay parallelizes across candidates; the 6.0\% cost edge is therefore a deployment-only figure.

\begin{table}[tbp]
\centering\small
\caption{\textbf{Realized search-inclusive cost per update} (mean over streams; budget cap 0.85M tokens/update for all search methods).}
\label{tab:cost}
\begin{tabular}{lcc}
\toprule
\textbf{Method} & Tokens/update & A100-hours/update \\
\midrule
\cmas{}      & 850k & 4.2 \\
AFlow-style  & 820k & 6.5 \\
MaAS-style   & 780k & 5.1 \\
GEPA-style   & 450k & 1.8 \\
Single agent & 50k  & 0.2 \\
\bottomrule
\end{tabular}
\end{table}

\textbf{Leakage and reproducibility protocol.} Freeze raw snapshots, release times, model versions, retrieval timestamps; lag features by actual publication availability with vintage macro data; date-controlled or counterfactual replay (never prompt-based memory suppression); strip real dates, event names, tickers in the controlled benchmark; disable web access and uncontrolled memory during replay; common random numbers for candidate/incumbent pairs; report factor exposures, residuals, costs, turnover, failed episodes; store diffs, traces, tool calls, approvals, rollbacks; pre-specify primary metrics, confidence methods, attack budgets, and final test intervals; release simulator, schema, prompts, and harness where licensing permits.

\section{Additional tables, figures, and ablations}
\label{app:extra}

\begin{figure}[t]
\centering
\includegraphics[width=0.72\textwidth]{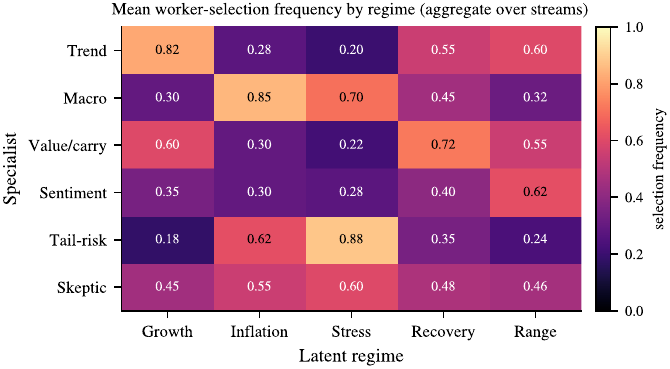}
\caption{\textbf{Selection vs.\ regime (RQ2).} Mean worker-selection frequency by latent regime over the 96 streams; rotation with the hidden regime (macro triples from Growth to Inflation, tail-risk peaks in Stress) is evidence consistent with regime-conditioned selection, and the shuffled-regime control (Table~\ref{tab:ablate}) removes it.}
\label{fig:dynamics}
\end{figure}

\begin{table}[t]
\centering
\caption{\textbf{Key ablations and controls} (Track~A held-out streams). The rows above the rule are one-component-removed ablations; the two rows below the rule are \emph{negative controls} run on full \cmas{}. \emph{Clone team size} is each variant's mean team size under the clone-collapse control; for the clone-control row itself this is by construction the same quantity as Full \cmas's entry (1.6), and its OOD utility of 0.28 matches the single agent, as expected under identical workers. 95\% hierarchical block-bootstrap CIs (clustered by regime schedule, seeds within) widen slightly relative to stream-level intervals (e.g., no-paired-evaluation OOD utility $[0.42,0.48]$), but every utility and false-promotion degradation relative to Full \cmas{} remains assessed by paired stream-level differences with Holm correction (full matrix omitted for space); the marginal CIs shown are descriptive, and the ``No parsimony'' and ``No adversarial tests'' false-promotion deltas are near-neutral.}
\label{tab:ablate}
\footnotesize
\begin{tabular}{lcccc}
\toprule
\textbf{Variant} & OOD utility $\uparrow$ & False promo.\ $\downarrow$ & Cost $\downarrow$ & Clone team size \\
\midrule
Full \cmas{}                    & \textbf{0.48} & \textbf{0.034} & 1.72 & \textbf{1.6} \\
No regime conditioning          & 0.44 & 0.040 & 1.70 & 3.2 \\
No paired evaluation            & 0.45 & 0.061 & 1.77 & 3.4 \\
No interventional credit        & 0.46 & 0.046 & 1.75 & 3.5 \\
No parsimony penalty            & 0.47 & 0.037 & 2.02 & 4.6 \\
No simultaneous correction      & 0.48 & 0.084 & 1.73 & 3.4 \\
No adversarial tests            & 0.47 & 0.036 & \textbf{1.69} & 3.3 \\
\midrule
Shuffled regimes (control)      & 0.40 & 0.038 & 1.71 & 3.3 \\
Identical-worker clones (control) & 0.28 & 0.033 & 1.46 & 1.6 \\
\bottomrule
\end{tabular}
\end{table}

\begin{figure}[tbp]
\centering
\input{fig_bench}
\caption{\textbf{\bench{} triangulation.} Track~A measures controlled meta-level capability against a computable surrogate dynamic oracle; Track~B provides leakage-controlled external validity on markets (18 replay windows); Track~C measures governance under a budgeted meta-adversary (576 episodes). All tracks feed a shared meta-level metric suite.}
\label{fig:bench}
\end{figure}

\begin{figure}[tbp]
\centering
\includegraphics[width=0.72\textwidth]{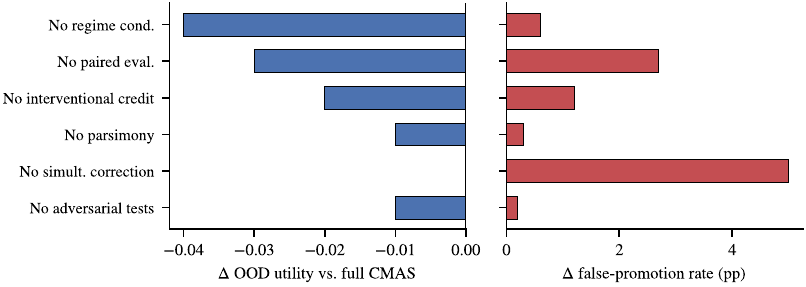}
\caption{\textbf{Ablation sensitivity} (visualizes Table~\ref{tab:ablate}). Regime conditioning, paired evaluation, and interventional credit dominate utility; paired evaluation and the simultaneous correction dominate promotion safety; the parsimony penalty prevents architecture bloat under clone-collapse.}
\label{fig:ablate}
\end{figure}

Table~\ref{tab:rq2} tabulates the RQ2 interventional-fidelity numbers quoted in the main text, and Table~\ref{tab:transfer} the RQ3 transfer numbers as stream-level means. The RQ6 governance suite is decomposed by attack family in Table~\ref{tab:gov}; Table~\ref{tab:ablate2} adds the proposer-value control (random/evolutionary candidates inside the \cmas{} evaluator+gate), and Figure~\ref{fig:ablate} visualizes the Table~\ref{tab:ablate} degradations.

\begin{algorithm}[h]
\caption{\cmas{} at outer-loop update $k$}
\label{alg:cmas}
\begin{algorithmic}[1]
\Require incumbent $h_k$, context $z_k$, archive $\mathcal{D}_k$, shadow window $W_k$, risk policy $\mathcal{C}$
\State retrieve diverse parent architectures from $\mathcal{D}_k$ via Eq.~\eqref{eq:retrieve}
\State meta-agent proposes typed candidates $\mathcal{H}_k$ with hypotheses and anticipated failure modes
\State compile candidates; reject invalid permissions, schemas, or graphs
\For{$c \in \mathcal{H}_k$}
  \State run paired shadow replay against $h_k$ (common random numbers)
  \State estimate the certified statistic $\Jc$ and full utility $J$, component credits $\phi$ (Eq.~\eqref{eq:credit}), uncertainty, and cost
  \State execute failure interventions (deployment-time eligibility checks; the benchmark-level negative controls of Section~\ref{sec:method} are run by the evaluator, not per candidate)
\EndFor
\State compute calibrated simultaneous scores of $\Jc$ (Eq.~\eqref{eq:score}) and the Pareto-feasible set
\If{some feasible $c$ has $\Score_k(c)>\epsilon$}
  \State canary-deploy the highest-score feasible candidate (ties broken toward fewer workers); retain rollback snapshot
\Else
  \State retain $h_k$
\EndIf
\State append traces, credits, failures, and deployment outcome to $\mathcal{D}_{k+1}$
\end{algorithmic}
\end{algorithm}

\begin{table}[tbp]
\centering\small
\caption{\textbf{RQ2 interventional fidelity} (Track~A). Architecture-selection accuracy is the fraction of outer-loop updates matching the architecture selected by the surrogate dynamic oracle at that update; credit $\rho$ is the rank correlation between estimated interventional credit and surrogate-model component value.}
\label{tab:rq2}
\begin{tabular}{lcc}
\toprule
\textbf{Method} & Selection accuracy $\uparrow$ & Credit $\rho$ $\uparrow$ \\
\midrule
Best adaptive baseline & 59.1\% & 0.32 \\
\textbf{\cmas{} (ours)} & \textbf{63.8\%} & \textbf{0.41} \\
\bottomrule
\end{tabular}
\end{table}

\begin{table}[tbp]
\centering
\caption{\textbf{Track~B external validity} ($N=18$ date-controlled replay windows; net of 10\,bps costs; ``best adaptive baseline'' is MaAS/AutoMaAS-style selection). This track tests whether Track~A conclusions survive contact with market data under leakage controls; with $N=18$ it is \emph{underpowered to separate adaptive methods}, differences are not tested for significance, and no row should be read as evidence of deployable alpha. Factor $\alpha$ is the residual after market/style attribution. Worker-compromise recovery is a governance quantity and is reported only in Table~\ref{tab:gov} (Track~C) to avoid conflating the two tracks. Because $N=18$, we report the sign pattern only: \cmas{} CEQ dispersion is $\pm1.1$\,pp (across-window standard deviation; mean $7.8\%$, min $6.8\%$, max $11.2\%$; the window distribution is right-skewed), so its worst windows overlap the baseline mean, and no cell should be read as a significance claim. Units: CEQ and factor $\alpha$ in annualized \%; Sharpe dimensionless; MDD and CVaR \%; turnover annualized fraction; violations fraction of decisions.}
\label{tab:replay}
\small
\setlength{\tabcolsep}{4.6pt}
\begin{tabular}{lccccccc}
\toprule
\textbf{Method} & CEQ\,\% $\uparrow$ & Sharpe $\uparrow$ & MDD $\downarrow$ & CVaR $\downarrow$ & $\alpha$\,\% $\uparrow$ & Turnover $\downarrow$ & Violations $\downarrow$ \\
\midrule
Risk parity          & 4.2 & 0.68 & 19.2 & 2.24 & 0.2 & 0.18 & 0.000 \\
Single agent + SC    & 5.1 & 0.84 & 17.8 & 2.05 & 0.9 & 0.31 & 0.034 \\
Fixed committee      & 5.8 & 0.95 & 18.4 & 2.12 & 1.2 & 0.44 & 0.049 \\
Rule router          & 6.7 & 1.04 & 15.6 & 1.88 & 1.6 & 0.32 & 0.026 \\
Best adaptive basel. & 7.4 & 1.12 & 14.5 & 1.79 & 2.0 & 0.41 & 0.031 \\
\midrule
\cmas{} (ours)       & 7.8 & 1.15 & 14.1 & 1.68 & 2.2 & 0.35 & 0.021 \\
\bottomrule
\end{tabular}
\end{table}

\begin{table}[tbp]
\centering\small
\caption{\textbf{Meta-level transfer} (Track~A held-out streams; retention $=$ held-out $/$ in-distribution utility). ``Best baseline'' is MaAS/AutoMaAS-style selection except in the backbone-switch row, where the relevant competitor is the search-based AFlow-style method (noted inline). The held-out-library baseline cell is ``--'' because MaAS-style selection cannot ingest new worker types without re-running search.}
\label{tab:transfer}
\setlength{\tabcolsep}{4.5pt}
\begin{tabular}{lccc}
\toprule
\textbf{Held-out factor} & \cmas{} & Best baseline & Relative \\
\midrule
Unseen regime mixtures        & 83.5\% & 78.5\% & $+5.0$\,pp \\
Held-out worker library (2 new) & 81.2\% & --     & recovers in 4 updates \\
Backbone switch (OOD util.\ cost) & 7.8\% & 11.8\% (AFlow-style) & $1.5\times$ smaller \\
Held-out violation rate (vs.\ 2.1\% ID) & 1.8\% & -- & safety clause met (no increase) \\
Held-out false promotion (vs.\ 3.4\% ID) & 4.1\% & -- & strict clause not met ($+0.7$\,pp) \\
\bottomrule
\end{tabular}
\end{table}

\begin{table}[tbp]
\centering\small
\caption{\textbf{Governance suite by attack family} (Track~C; 576 episodes nested in 96 streams; recovery within three updates). Aggregate Wilson 95\% interval $[0.75, 0.81]$; per-family Wilson and cluster-by-stream intervals are omitted for space. Five single-attack families plus one multi-attack combination; the aggregate spans all six scenarios. Aggregate rollback precision/recall $0.85/0.80$.}
\label{tab:gov}
\begin{tabular}{lcccc}
\toprule
\textbf{Attack family} & \cmas{} recovery & ($k/n$) & Best baseline & Committee \\
\midrule
Instruction-corrupted worker & 0.81 & 78/96 & 0.74 & 0.54 \\
Anti-correlated worker        & 0.79 & 76/96 & 0.72 & 0.53 \\
Stale/missing data            & 0.77 & 74/96 & 0.70 & 0.51 \\
Inference-cost shock          & 0.76 & 73/96 & 0.69 & 0.50 \\
Worker dropout                & 0.80 & 77/96 & 0.73 & 0.52 \\
\midrule
Multi-attack combination      & 0.75 & 72/96 & 0.68 & 0.49 \\
\midrule
\midrule
Aggregate                     & 0.78 & 450/576 & 0.71 & 0.52 \\
\bottomrule
\end{tabular}
\end{table}

\begin{figure}[tbp]
\centering
\includegraphics[width=0.44\textwidth]{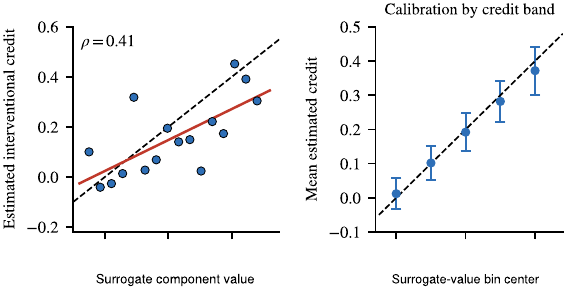}
\caption{\textbf{Credit fidelity (RQ2).} Estimated interventional credit shows moderate positive rank agreement with surrogate-model component value ($\rho{=}0.41$, $n{=}16$ component types pooled across streams; regression line shown), strongest among high-credit components. The pooled scatter is illustrative only; the statistic we rely on is the \emph{per-stream} distribution of $\rho$ over the 96 streams: median $\rho=0.41$, 95\% CI $[0.35,0.47]$, excluding zero.}
\label{fig:credit}
\end{figure}

\section{Example update trace}
\label{app:trace}
One end-to-end update from the promotion logs: the proposer submits the diff \{add Tail-Risk, remove Sentiment\} with hypothesis and expected mechanism; the compiler validates; a 26-week paired shadow replay yields $\Score_k=0.028>\epsilon=0.02$ with all eligibility filters passing; the candidate is promoted and enters canary deployment; no rollback trigger fires over the subsequent four weeks. The full trace (prompts, per-block deltas, credit vector, trigger telemetry) is omitted for space.

\section{Failure modes and negative results}
\label{app:fail}
\cmas{} is not uniformly best. It underperforms the static rule router in 14 of 96 streams, predominantly during prolonged \emph{Range} regimes, where switching costs and exploration penalties outweigh the benefit of adaptation and a fixed cheap policy is hard to beat. The gate is not infallible in either direction. In Stream~42, a tail-risk specialist was promoted immediately before a sudden volatility crush. The canary-deployment rollback trigger fired, but only after a 1.2\% simulated canary-path drawdown was incurred, which is the concrete instance behind the pre-rollback-harm limitation in Section~\ref{sec:discussion}. Conversely, in Stream~11 a sparse architecture that the oracle ranks highly was rejected because its calibrated score missed the $\epsilon=0.02$ margin by 0.002 under high shadow-window variance, an example of the conservatism cost quantified in RQ4 (42\% retention). Per-stream loss maps and the full promotion/rejection logs for these cases are omitted for space.

\section{Reproducibility}
\label{app:repro}

This appendix is the source of record for the configuration that produced the reported results. A reproducibility package (simulator, environment interface, baseline adapters, evaluation scripts, seed manifests, frozen configurations, one full worker/aggregation prompt, and ten fully traced streams) is in preparation for public release. Licensed datasets and hosted-model credentials cannot be redistributed, and scripts and manifests document the access procedure. Failed worker calls are retried up to three times (retries counted in the search-inclusive budget); after three failures the worker returns a neutral no-opinion tuple, and wasted tokens (2.1\% of search tokens on average) are logged. Settings are mirrored in \texttt{config\_frozen.json}.

\begin{table}[tbp]
\centering\small
\caption{\textbf{Information chronology at update $k$.} This distinguishes the overlapping protocol from the sealed rerun and makes the proposal/evaluation boundary explicit.}
\label{tab:timeline}
\begin{tabular}{L{1.0cm}L{3.0cm}L{8.5cm}}
\toprule
\textbf{Step} & \textbf{Time / object} & \textbf{Information boundary} \\
\midrule
1 & archive cutoff $t_k$ & build $\mathcal D_k$ only from outcomes and credits released before $t_k$ \\
2 & candidate generation & freeze $\mathcal H_k$; no access to outcomes in the designated $W_k$ \\
3 & paired replay $W_k$ & run candidate and incumbent on identical packets, shocks, seeds, and tool responses \\
4 & gate decision & apply eligibility filters and $\Score_k$; retain incumbent or enter simulated canary stage \\
5 & canary interval & candidate has limited simulated authority alongside the incumbent until the next update; rollback restores the incumbent architecture and state snapshot \\
6 & archive release & replay labels, canary outcomes, and credits become available only to $\mathcal D_{k+1}$ \\
\midrule
Overlap & overlapping protocol & successive 26-week $W_k$ overlap at monthly cadence; a later pool can use earlier released outcomes, but no pool sees its own $W_k$ before freezing \\
Sealed check & calibration rerun & fresh, non-overlapping paths per update; same frozen score and margin, no retuning \\
Final test & reported results & untouched during search/calibration; only final frozen configuration is scored \\
\bottomrule
\end{tabular}
\end{table}

\begin{table}[tbp]
\centering\small
\caption{\textbf{Calibration and evaluation stages.} The source records the roles and frozen choices below; it does not contain a public preregistration timestamp or anonymous commit hash, so none is claimed.}
\label{tab:splits}
\begin{tabular}{L{2.8cm}L{3.9cm}L{5.1cm}}
\toprule
\textbf{Stage} & \textbf{Used for} & \textbf{Frozen before subsequent stage} \\
\midrule
Development/training & proposer/archive development; utility normalizers & search space, typed edit interface, normalizers \\
Validation/calibration & score form and sensitivity analysis & block length 4, 99th-percentile winsorization, $\epsilon=0.02$, candidate pool 8, rollback thresholds \\
Final Track A test & 96 streams; 2,112 update decisions & no tuning or candidate-pool adaptation to final outcomes \\
Sealed rerun & non-overlapping empirical calibration check & reuses the same score, margin, and eligibility rules \\
\bottomrule
\end{tabular}
\end{table}

\textbf{Denominator transparency.} The main text reports both false promotions/all updates (3.4\%) and false promotions/actual promotions (5.86\%), plus full-$J$ reduction (1.2\%), negative-$\Jc$ promotion (0.4\%), and streams with at least one false promotion (11.4\%). The manuscript source contains rounded rates rather than raw integer event logs; rounded-rate equivalents are approximately 72 false promotions among 2,112 updates, approximately 1,225 promotions, approximately 25 full-$J$ reductions, approximately 8 negative-$\Jc$ promotions, and approximately 11 affected streams. These are explicitly approximate; exact integer logs are part of the planned reproducibility release.

\textbf{Metric details.} Adaptation delay is right-censored at eight post-transition updates; a static method can therefore receive a finite capped delay. Selection accuracy uses exact equality with the surrogate-optimal architecture, so functionally equivalent or near-optimal graphs count as mismatches; normalized regret is the primary metric. OOD normalization uses frozen training minima/maxima without clipping test values.

\textbf{Per-decision Wilson intervals (secondary).} Treating the $2{,}112$ update-level promotion decisions as independent, the Wilson interval for observed false promotion is $[2.7,4.3]\%$ at the 5\% target vs.\ $[7.3,9.6]\%$ without the simultaneous correction; these intervals ignore within-stream and within-schedule clustering and are secondary to the schedule-clustered estimates reported in the main text.

\textbf{Seed separation and planned reproducibility release} (referenced from Section~\ref{sec:protocol}). Seeds for sampling, simulation, and proposal are separated; an open-weight track reduces API dependence; failed runs, costs, prompts, versions, and traces are logged in that package.

\textbf{Configuration (stated in-paper).} Architecture space and promotion rule are pre-specified in Appendix~\ref{app:space}: DAG with $\le 10$ edges; any subset of six worker roles (per-role backbone a separate field); budgets $\{1\times,2\times,4\times\}$; five aggregation and five verification options; one typed field edited per candidate.

Statistics: 96 streams (12 seeds $\times$ 8 regime schedules), ${\sim}122$-week streams (26-week unscored burn-in; 96 scored weekly decisions) with monthly updates ($K\approx22$), 95\% circular block-bootstrap CIs with pre-selected block length, Holm correction \emph{across the seven inferential Table~\ref{tab:main} comparisons} (average team size is descriptive and excluded) for \cmas{} vs.\ best baseline, common random numbers for candidate/incumbent pairs, separated seeds for sampling/simulation/proposal. We apply the budget \emph{per update}: $\delta=0.05$ is the family-wise target across the $|\mathcal H_k|=8$ candidates evaluated in one update (per-candidate level $\delta/|\mathcal H_k|\approx0.006$). The anytime-valid schedule of Appendix~\ref{app:proof} would additionally control error across the open-ended update sequence, but, as scoped in Section~\ref{sec:method}, it was \emph{not activated} for the reported Track~A runs; the per-stream family-wise rate of RQ4 should be read accordingly. The reported 3.4\% is the per-update false-promotion rate over $2{,}112$ decisions ($96\times K$), directly comparable to the 5\% per-update target (Figure~\ref{fig:gate}).

\textbf{Normalization of Eq.~\eqref{eq:utility}.} Every penalty term is mapped to a common scale before the $\lambda$ weights are applied: per-term min--max normalization over the training interval, with the normalizers frozen at deployment (no test-time re-normalization). As a local scale illustration only, holding all other normalized penalty terms fixed, OOD utility $0.44\to0.48$ maps to approximately $4.12\%\to4.58\%$ annualized CEQ; this is not an independent economic-performance claim.

\begin{table}[tbp]
\centering\small
\caption{\textbf{Configuration used for the reported results.} These values are the source of record for the reported results. Proposer and red-team/critic model: Llama-3-70B-Instruct (cutoff Dec 2023; temperature 0.2, top-$p$ 0.9); candidate utilities are computed by deterministic replay code. Workers: Mixtral-8x7B-Instruct-v0.1 (vendor cutoff undisclosed; replay restricted to post--January 2024 data; temperature 0.4) and Llama-3-8B-Instruct (cutoff Mar 2023; temperature 0.3). Hosted reasoning model (backbone-transfer test): Claude 3 Opus via API (cutoff Aug 2023; temperature 0.1). Seeds: 42 (sampling), 1337 (simulation), 2024 (proposal).}
\label{tab:hparams}
\small
\setlength{\tabcolsep}{4pt}
\begin{tabular}{L{2.6cm}L{6.6cm}L{3.6cm}}
\toprule
Symbol & Meaning & Value \\
\midrule
$|\mathcal H_k|$ & candidate-pool size per update & 8 \\
$K$ & outer-loop updates per stream & $\approx$22 (monthly) \\
$\epsilon$ & promotion margin (normalized utility) & 0.02 \\
$\delta$ & family-wise false-promotion budget (per update) & 0.05 \\
$W_k$ & paired replay-window length & 26 weeks \\
block length & circular block bootstrap & 4 weeks \\
Shapley budget & adaptive (screen all; full on top-3) & 8 screen; 24$\times$3 \\
$\gamma$ & CEQ risk aversion & 2.0 \\
$\lambda_{\mathrm{cvar}},\lambda_{\mathrm{dd}},\lambda_{\mathrm{cost}}$ & risk/cost weights & 0.5, 0.3, 0.2 \\
$\lambda_{\mathrm{switch}},\lambda_{\mathrm{turn}},\lambda_{\mathrm{viol}}$ & switch/turnover/violation weights & 0.15, 0.1, 1.0 \\
retrieved parents & per proposal & 4 \\
$\beta$ & retrieval weights (sim, utility, unc., novelty); Pareto prefilter first & 0.4, 0.3, 0.2, 0.1 \\
$\ell$ & regime-kernel bandwidth (standardized $z_k$) & 0.35 \\
$(\omega_V,\omega_E,\omega_b,\omega_\rho)$ & worker/graph/budget/typed-field distance weights, Eq.~\eqref{eq:dist} & 0.3, 0.4, 0.1, 0.2 \\
$\epsilon_{\mathrm{delay}}$ & adaptation-delay tolerance (Section~\ref{sec:protocol}) & 0.05 \\
rollback thresholds & canary MDD / CVaR violation / utility drift & $>$8\%; $>2\times$ baseline; $>$0.03 \\
LinUCB & menu size; exploration $\alpha$ & 15; 1.0 \\
$M$ & raw-utility-scale range bound (paired deltas concentrate to ${\approx}0.04$) & 0.15 \\
$B_{\mathrm{eff}}$ & effective independent blocks (median) & 5.8 \\
streams & seeds $\times$ schedules & 96 ($12\times8$) \\
gate decisions & total across streams & 2{,}112 \\
tokens/update & staged Shapley $+$ 8B routing & 0.85M \\
total compute & tokens; aggregate A100-hours (4.2/update, Table~\ref{tab:cost}) & 1.79B; ${\approx}8{,}870$ \\
$\xi_k$ & replay-to-deployment margin (Track~A) & 0 \\
worker model pool & small / strong open-weight; hosted transfer model & 3 models (proposer/critic separate) \\
\bottomrule
\end{tabular}
\end{table}

\textbf{Worked packet (Track~A).} A ``macro'' worker at week $t$ receives a packet of the form
\{\texttt{context}: $z_k$ descriptor; \texttt{signal}: rendered $y_{i,t}$ as bucketed factor tilts with a stated noise level; \texttt{history}: last-4-week own scores and realized outcomes; \texttt{role}: ``estimate 4-week factor direction and size, flag regime risk''\} and returns $o_{i,t}=$ \{\texttt{scores}: per-asset tilts; \texttt{confidence}: 0--1 calibrated; \texttt{uncertainty}: interval; \texttt{evidence\_ids}; \texttt{risk\_flags}\}. The simulator sets signal quality via $H,G$; the LLM contributes calibration, aggregation-relevant disagreement, and failure modes.

\textbf{Cost accounting.} Per-method realized usage is summarized in Table~\ref{tab:cost}. Structurally, \cmas's search (paired replay $+$ Shapley) is its dominant cost, and its total exceeds every compared method's, so break-even against the Table~\ref{tab:main} deployment-only edge depends on the per-update deployment horizon.

\section{Extended responsible use and governance}
\label{app:responsible}

This appendix expands the statement in Section~\ref{sec:discussion}. Nothing here is an
evaluated result: it specifies the governance envelope the method is designed to sit inside,
and separates what we implemented from what a real operator would still have to supply.

\textbf{Implemented in this work.} The deterministic risk projection $\Pi_{\mathcal C_t}$,
authorization layer, accounting, and data cutoffs are outside the LLM boundary and cannot be
edited by any proposal (Figure~\ref{fig:overview}). Hard authorization, evidence, leakage, and
risk checks are \emph{eligibility filters} applied before the statistical gate, so a candidate
failing them is rejected regardless of measured utility. The candidate space is closed
(Table~\ref{tab:space}) and every promotion is logged as a typed diff with hypothesis,
replay evidence, credit vector, and outcome.

\textbf{Specified but not evaluated.} Human sign-off, emergency halt, incident review, and
audit-log ownership are design requirements, not measured quantities. We recommend that the
party setting the objective, margin $\epsilon$, risk limits, and rollback triggers be distinct
from the party building the proposer, and that promotion authority and halt authority rest with
a reviewer independent of both. The canary stage in this paper is \emph{simulated}: a real
limited-authority deployment would need separate approval, position and blast-radius limits,
and an independent monitoring path.

\textbf{Failure modes of the oversight itself.} Oversight can fail in four ways. Automation bias: a
reviewer presented with a passing score and a plausible hypothesis may approve without
inspecting the diff. Rubber-stamping turns routine sign-off into a formality, and is
likelier the higher the promotion rate, which is one reason the 42\% incumbent-retention
rate matters. Evaluator misspecification: if the objective is wrong, the gate faithfully certifies
the wrong thing, which is exactly what the return-only probe of RQ7 demonstrates. Specification
error in the immutable layer completes the list; constraints protect only if correctly written,
and nothing in \cmas{} detects a wrong constraint.

\textbf{Misuse.} A meta-agent search loop is objective-agnostic. The same machinery could
optimize agent fleets for surveillance, manipulation, or exploitative coordination; scale a
harmful objective more cheaply; search around an imperfect safety evaluator, since a gate that
screens on a measured statistic will also find architectures that satisfy the statistic without
satisfying its intent; or produce architectures whose complexity obscures accountability. The
parsimony tie-break and the bounded edit set mitigate the last of these, but do not address the
first three, which are governance rather than technical problems.

\textbf{Residual risk.} A candidate can pass the gate and reduce full $J$
(1.2\% of decisions here). Sequence-level error is uncontrolled and the per-stream bound implied
by the reported configuration is vacuous (Appendix~\ref{app:proof}). Canary-stage harm can occur
before a trigger fires, as in Stream~42 (Appendix~\ref{app:fail}). The simulator may omit
correlated failure modes that matter in deployment. Track~B is underpowered and is not evidence
of deployable alpha. Brokerage connectivity and live-trading validation are excluded entirely,
and a real deployment would require validation by parties independent of the developers.

\section{Extended related work and positioning}
\label{app:related}

On the optimization side, DSPy compiles declarative pipelines and GEPA evolves prompts reflectively, while Trace generalizes backpropagation-style credit to computational graphs~\cite{dspy,gepa,trace}; Meta-Harness searches full harness code with prior traces~\cite{metaharness}.

Concurrent 2026 work studies gated harness evolution separating LLM proposals from deterministic crediting~\cite{gatedqd}, automated harness optimization~\cite{harbor}, online adaptation for self-improving agents~\cite{contharness}, robust agentic-system learning~\cite{autoras}, and sustained construction/routing over open-ended non-stationary streams~\cite{adaptiveharness}; \cmas{} differs in targeting repeated candidate--incumbent \emph{replacement} under non-stationarity, with paired incumbent comparison, rollback, and false-promotion measurement first-class (Tables~\ref{tab:compare-search}--\ref{tab:compare-safety}).

\textbf{Safe improvement, conservative exploration, and anytime-valid inference.} Our gate belongs to the tradition of high-confidence policy improvement~\cite{hcpi}, safe policy improvement with baseline bootstrapping~\cite{spibb}, and conservative bandits/exploration bounding performance below a baseline while learning~\cite{consbandit}; the surrogate dynamic-oracle-regret metric follows non-stationary and switching-cost online learning~\cite{besbes,herbster}; our paired shadow replay is related to off-policy evaluation with confidence intervals~\cite{ope}; it pairs candidate and incumbent under common random numbers rather than reweighting a logged policy. Paired replay is available because the simulator and counterfactual replay permit \emph{re-execution} of candidate and incumbent on the same shocks, the step that importance-weighted, doubly robust, and fitted-$Q$ estimators are designed to make unnecessary when only logged data are available. Three properties are new relative to this literature: (a)~the action space is \emph{typed architectures} with structure and a distance (Eq.~\eqref{eq:dist}), not a fixed policy class; (b)~candidates are generated \emph{adaptively by an LLM proposer}, breaking i.i.d.\ candidate assumptions, which is why the guarantee is conditional on the pool; and (c)~evaluation uses LLM execution traces with interventional credit. For open-ended deployment, $\delta/K$ can be replaced by an anytime-valid schedule or confidence sequence~\cite{anytime} (Appendix~\ref{app:proof}).

\textbf{Safety, verification, and meta-level failure.} MaMa casts safe agent-system design as a game against a worker-compromising meta-adversary~\cite{mama}; HarnessAudit audits full trajectories for boundary and information-flow violations~\cite{harnessaudit}, SafeHarness integrates lifecycle verification and rollback~\cite{safeharness}, and other work verifies agent graphs and attributes failures structurally~\cite{verifmeta}. The Meta-Agent Challenge argues that task benchmarks do not measure the ability to \emph{develop} agent systems~\cite{machallenge}. \cmas{} operationalizes these via false-promotion measurement, worker-compromise tests, trace-level credit, canary deployment, and rollback.

\textbf{Financial agents and evaluation validity.} Financial systems include memory-based agents~\cite{finmem}, trading-firm committees~\cite{tradingagents}, prompt adaptation from market feedback~\cite{atlas}, fine-grained expert teams~\cite{expertteams}, and self-driving portfolios~\cite{selfdriving}. Leakage-controlled studies show LLM backtests are inflated by parametric memory and passive factor exposure~\cite{profitmirage,lookahead,livebench,memctrl}; we therefore make the simulator the primary controlled ground-truth test and treat market replay as \emph{external validity} only.


\end{document}

%% file: icons.tex
\definecolor{coral}{HTML}{D96B6B}
\definecolor{amber}{HTML}{E9A83A}
\definecolor{leaf}{HTML}{5FA04A}
\definecolor{sky}{HTML}{5B8FCf}
\definecolor{lavender}{HTML}{8E7BC4}
\definecolor{sea}{HTML}{3F9C9E}
\definecolor{ink}{HTML}{3B3B3B}
\definecolor{paperbg}{HTML}{ECE8E2}
\definecolor{softblue}{HTML}{DCE7F5}
\definecolor{softgreen}{HTML}{DEEAD6}
\definecolor{softpurple}{HTML}{E8E1F2}
\definecolor{softteal}{HTML}{D6ECEC}
\definecolor{softred}{HTML}{F6DEDE}
\definecolor{softamber}{HTML}{F8EDD1}
\definecolor{softgray}{HTML}{E9E7E3}

\tikzset{
  globe/.pic={
    \draw[line width=0.55pt] (0,0) circle (0.19);
    \draw[line width=0.4pt] (0,0) ellipse (0.082 and 0.19);
    \draw[line width=0.4pt] (-0.19,0) -- (0.19,0);
    \draw[line width=0.4pt] (-0.16,0.095) -- (0.16,0.095);
    \draw[line width=0.4pt] (-0.16,-0.095) -- (0.16,-0.095);
  },
  mag/.pic={
    \draw[line width=0.9pt] (0,0) circle (0.13);
    \draw[line width=1.3pt] (0.092,-0.092) -- (0.20,-0.20);
  },
  agent/.pic={
    \draw[line width=0.7pt] (0,0.08) circle (0.09);
    \draw[line width=0.7pt] (-0.15,-0.19) arc (208:-28:0.165);
  },
  db/.pic={
    \draw[line width=0.7pt] (-0.16,0.16) -- (-0.16,-0.16) arc (180:360:0.16 and 0.06) -- (0.16,0.16);
    \draw[line width=0.7pt] (0,0.16) ellipse (0.16 and 0.06);
    \draw[line width=0.4pt] (-0.16,0.03) arc (180:360:0.16 and 0.06);
    \draw[line width=0.4pt] (-0.16,-0.07) arc (180:360:0.16 and 0.06);
  },
  shield/.pic={
    \draw[line width=0.8pt] (0,0.20) -- (0.16,0.12) -- (0.16,-0.04)
      .. controls (0.16,-0.15) and (0.07,-0.20) .. (0,-0.23)
      .. controls (-0.07,-0.20) and (-0.16,-0.15) .. (-0.16,-0.04)
      -- (-0.16,0.12) -- cycle;
    \draw[line width=1.0pt] (-0.07,0.0) -- (-0.02,-0.07) -- (0.08,0.08);
  },
  chk/.pic={ \draw[line width=1.2pt] (-0.08,0.0) -- (-0.02,-0.08) -- (0.10,0.09); },
  crs/.pic={ \draw[line width=1.2pt] (-0.08,0.08) -- (0.08,-0.08);
             \draw[line width=1.2pt] (-0.08,-0.08) -- (0.08,0.08); },
  gear/.pic={
    \draw[line width=0.7pt] (0,0) circle (0.12);
    \draw[line width=0.7pt] (0,0) circle (0.055);
    \foreach \a in {0,45,...,315}{ \draw[line width=1.9pt] (\a:0.12) -- (\a:0.175); }
  },
  lock/.pic={
    \draw[line width=0.8pt, rounded corners=1pt] (-0.13,-0.02) rectangle (0.13,-0.22);
    \draw[line width=0.8pt] (-0.08,-0.02) -- (-0.08,0.06) arc (180:0:0.08) -- (0.08,-0.02);
    \fill (0,-0.11) circle (0.018);
  },
  codei/.pic={
    \draw[line width=1.0pt] (-0.06,0.11) -- (-0.17,0.0) -- (-0.06,-0.11);
    \draw[line width=1.0pt] (0.06,0.11) -- (0.17,0.0) -- (0.06,-0.11);
    \draw[line width=1.0pt] (0.045,0.13) -- (-0.045,-0.13);
  },
  refresh/.pic={
    \draw[line width=1.0pt, -{Stealth[length=1.6mm]}] (0.16,0.03) arc (0:295:0.16);
  },
  uptrend/.pic={
    \draw[line width=1.0pt, -{Stealth[length=1.4mm]}] (-0.19,-0.08) -- (-0.06,0.03) -- (0.03,-0.04) -- (0.19,0.14);
  },
  downarr/.pic={
    \draw[line width=1.0pt, -{Stealth[length=1.4mm]}] (-0.19,0.10) -- (-0.06,-0.02) -- (0.03,0.04) -- (0.19,-0.16);
  },
  warn/.pic={
    \draw[line width=0.75pt, line join=round] (0,0.18) -- (0.18,-0.15) -- (-0.18,-0.15) -- cycle;
    \draw[line width=1.0pt] (0,0.10) -- (0,-0.04);
    \fill (0,-0.10) circle (0.018);
  },
  coins/.pic={
    \draw[line width=0.6pt] (0,-0.08) ellipse (0.16 and 0.06);
    \draw[line width=0.6pt] (-0.16,-0.08) -- (-0.16,0.0) arc(180:360:0.16 and 0.06) -- (0.16,0.0);
    \draw[line width=0.6pt] (0,0.0) ellipse (0.16 and 0.06);
    \draw[line width=0.4pt] (0,-0.04) ellipse (0.16 and 0.06);
  },
  balance/.pic={
    \draw[line width=0.85pt] (0,0.20) -- (0,-0.17);
    \draw[line width=0.85pt] (-0.19,0.15) -- (0.19,0.15);
    \draw[line width=0.55pt] (-0.19,0.15) -- (-0.25,0.03) -- (-0.13,0.03) -- cycle;
    \draw[line width=0.55pt] (0.19,0.15) -- (0.25,0.03) -- (0.13,0.03) -- cycle;
    \draw[line width=0.85pt] (-0.09,-0.17) -- (0.09,-0.17);
  },
  bulb/.pic={
    \draw[line width=0.85pt] (0,0.04) circle (0.13);
    \draw[line width=0.85pt] (-0.06,-0.10) -- (-0.06,-0.17) -- (0.06,-0.17) -- (0.06,-0.10);
    \draw[line width=0.6pt] (-0.05,-0.135) -- (0.05,-0.135);
    \draw[line width=0.9pt] (0,0.19) -- (0,0.24);
    \draw[line width=0.9pt] (0.14,0.12) -- (0.18,0.16);
    \draw[line width=0.9pt] (-0.14,0.12) -- (-0.18,0.16);
  },
  packet/.pic={
    \draw[line width=0.8pt, rounded corners=1pt] (-0.19,-0.13) rectangle (0.19,0.13);
    \draw[line width=0.6pt] (-0.19,0.09) -- (0,-0.02) -- (0.19,0.09);
  },
  clock/.pic={
    \draw[line width=0.8pt] (0,0) circle (0.16);
    \draw[line width=0.9pt] (0,0) -- (0,0.10);
    \draw[line width=0.9pt] (0,0) -- (0.08,-0.02);
  },
  brokenloop/.pic={
    \draw[line width=0.9pt, dash pattern=on 1.4pt off 1.2pt] (0,0) circle (0.15);
    \draw[line width=1.2pt] (0.12,0.12) -- (-0.12,-0.12);
  },
  rocket/.pic={
    \draw[line width=0.8pt] (0,0.18) .. controls (0.11,0.05) and (0.11,-0.05) .. (0.06,-0.13)
      -- (-0.06,-0.13) .. controls (-0.11,-0.05) and (-0.11,0.05) .. (0,0.18);
    \draw[line width=0.6pt] (0,0.06) circle (0.045);
    \draw[line width=0.7pt] (-0.06,-0.10) -- (-0.13,-0.17) -- (-0.05,-0.13);
    \draw[line width=0.7pt] (0.06,-0.10) -- (0.13,-0.17) -- (0.05,-0.13);
  },
}

\tikzset{
  star5/.style={star, star points=5, star point ratio=0.48, draw=amber!75!black,
                fill=amber, line width=0.35pt, inner sep=0pt, minimum size=#1},
  star5/.default=9pt,
  xnode/.style={circle, draw=coral!70!black, fill=coral, text=white, inner sep=0pt,
                minimum size=8pt, font=\fontsize{5.5}{5.5}\bfseries\selectfont},
  cnode/.style={circle, draw=leaf!60!black, fill=leaf, text=white, inner sep=0pt,
                minimum size=8pt, font=\fontsize{5.5}{5.5}\bfseries\selectfont},
}

\tikzset{
  robot/.pic={
    \draw[line width=0.7pt] (0,0.30) -- (0,0.21);
    \fill (0,0.325) circle (0.026);
    \draw[line width=0.85pt, rounded corners=2.2pt] (-0.17,0.21) rectangle (0.17,-0.07);
    \fill (-0.075,0.10) circle (0.03);
    \fill (0.075,0.10) circle (0.03);
    \draw[line width=0.65pt] (-0.06,0.008) -- (0.06,0.008);
    \draw[line width=0.75pt] (-0.17,0.115) -- (-0.215,0.115);
    \draw[line width=0.75pt] (0.17,0.115) -- (0.215,0.115);
    \draw[line width=0.85pt, rounded corners=1.6pt] (-0.13,-0.10) rectangle (0.13,-0.26);
  },
  rtrend/.pic={\draw[line width=0.9pt,-{Stealth[length=0.9mm]}] (-0.09,-0.05)--(-0.02,0.02)--(0.02,-0.02)--(0.10,0.08);},
  rglobe/.pic={\draw[line width=0.5pt] (0,0) circle (0.09);\draw[line width=0.35pt] (0,0) ellipse (0.04 and 0.09);\draw[line width=0.35pt] (-0.09,0)--(0.09,0);\draw[line width=0.3pt](-0.075,0.045)--(0.075,0.045)(-0.075,-0.045)--(0.075,-0.045);},
  rumbrella/.pic={\draw[line width=0.6pt] (0,0.03)--(0,-0.10);\draw[line width=0.6pt] (0,-0.10) arc (0:-110:0.035);\draw[line width=0.7pt] (-0.11,0.03) arc (180:0:0.11 and 0.075);\draw[line width=0.45pt] (-0.11,0.03)--(0.11,0.03);},
  rq/.pic={\node[font=\fontsize{7}{7}\bfseries\selectfont]{?};},
  flask/.pic={
    \draw[line width=0.85pt] (-0.055,0.22) -- (-0.055,0.03) -- (-0.17,-0.19)
      .. controls (-0.19,-0.25) and (0.19,-0.25) .. (0.17,-0.19) -- (0.055,0.03) -- (0.055,0.22);
    \draw[line width=0.75pt] (-0.10,0.22) -- (0.10,0.22);
    \fill[fill opacity=0.22] (-0.135,-0.11) -- (-0.17,-0.19) .. controls (-0.19,-0.25) and (0.19,-0.25) .. (0.17,-0.19) -- (0.135,-0.11) -- cycle;
    \fill (-0.05,-0.15) circle (0.014)(0.05,-0.17) circle (0.014)(0.10,-0.12) circle (0.012);
  },
  candles/.pic={
    \draw[line width=0.8pt] (-0.17,-0.14) -- (0.19,-0.14);
    \draw[line width=0.6pt] (-0.10,-0.04) -- (-0.10,0.16);\draw[line width=2.4pt] (-0.10,0.0) -- (-0.10,0.11);
    \draw[line width=0.6pt] (0.0,-0.10) -- (0.0,0.10);\draw[line width=2.4pt] (0.0,-0.07) -- (0.0,0.05);
    \draw[line width=0.6pt] (0.10,0.0) -- (0.10,0.20);\draw[line width=2.4pt] (0.10,0.05) -- (0.10,0.16);
  },
  bug/.pic={
    \draw[line width=0.85pt] (0,0.02) ellipse (0.10 and 0.15);
    \draw[line width=0.55pt] (-0.04,0.16) -- (-0.08,0.25)(0.04,0.16) -- (0.08,0.25);
    \fill (-0.08,0.25) circle (0.017)(0.08,0.25) circle (0.017);
    \draw[line width=0.55pt] (-0.10,0.07) -- (-0.21,0.13)(-0.11,-0.02) -- (-0.22,-0.02)(-0.10,-0.11) -- (-0.20,-0.17);
    \draw[line width=0.55pt] (0.10,0.07) -- (0.21,0.13)(0.11,-0.02) -- (0.22,-0.02)(0.10,-0.11) -- (0.20,-0.17);
    \draw[line width=0.5pt] (0,-0.13) -- (0,0.17);
  },
  gauge/.pic={
    \draw[line width=1.0pt] (-0.20,-0.02) arc (180:0:0.20);
    \draw[line width=1.0pt] (-0.20,-0.02) -- (0.20,-0.02);
    \foreach \a in {24,72,108,156}{\draw[line width=0.5pt] (\a:0.20) -- (\a:0.16);}
    \draw[line width=1.1pt, -{Stealth[length=1.1mm]}] (0,-0.02) -- (0.11,0.14);
    \fill (0,-0.02) circle (0.022);
  },
  hmm/.pic={
    \foreach \i in {0,1,2}{\draw[line width=0.6pt] (\i*0.16,0) circle (0.045);}
    \draw[line width=0.6pt,-{Stealth[length=0.8mm]}] (0.05,0.02) arc (150:30:0.05);
    \draw[line width=0.6pt,-{Stealth[length=0.8mm]}] (0.21,0.02) arc (150:30:0.05);
  },
}

\tikzset{
  rcoin/.pic={\draw[line width=0.6pt] (0,0) circle (0.095);\draw[line width=0.5pt] (0,-0.05)--(0,0.05);\draw[line width=0.45pt](-0.028,0.028)--(0.028,0.028)(-0.028,-0.028)--(0.028,-0.028);},
  rbubble/.pic={\draw[line width=0.6pt, rounded corners=1.5pt] (-0.10,-0.01) rectangle (0.10,0.13);\draw[line width=0.6pt] (-0.04,-0.01) -- (-0.07,-0.08) -- (0.01,-0.01);\fill (-0.05,0.06) circle (0.009)(0,0.06) circle (0.009)(0.05,0.06) circle (0.009);},
  cup/.pic={\draw[line width=0.45pt] (-0.13,-0.05)--(-0.13,0.05);\draw[line width=1.7pt] (-0.13,-0.025)--(-0.13,0.025);
            \draw[line width=0.45pt] (0,-0.01)--(0,0.13);\draw[line width=1.7pt] (0,0.02)--(0,0.09);
            \draw[line width=0.45pt] (0.13,0.03)--(0.13,0.19);\draw[line width=1.7pt] (0.13,0.07)--(0.13,0.15);},
  cdn/.pic={\draw[line width=0.45pt] (-0.13,0.03)--(-0.13,0.19);\draw[line width=1.7pt] (-0.13,0.07)--(-0.13,0.15);
            \draw[line width=0.45pt] (0,-0.01)--(0,0.13);\draw[line width=1.7pt] (0,0.02)--(0,0.09);
            \draw[line width=0.45pt] (0.13,-0.05)--(0.13,0.05);\draw[line width=1.7pt] (0.13,-0.025)--(0.13,0.025);},
  cv/.pic={\draw[line width=0.45pt] (-0.13,0.02)--(-0.13,0.16);\draw[line width=1.7pt] (-0.13,0.05)--(-0.13,0.12);
           \draw[line width=0.45pt] (0,-0.06)--(0,0.06);\draw[line width=1.7pt] (0,-0.03)--(0,0.03);
           \draw[line width=0.45pt] (0.13,0.02)--(0.13,0.16);\draw[line width=1.7pt] (0.13,0.05)--(0.13,0.12);},
  cfl/.pic={\draw[line width=0.45pt] (-0.13,0.0)--(-0.13,0.12);\draw[line width=1.7pt] (-0.13,0.03)--(-0.13,0.09);
            \draw[line width=0.45pt] (0,0.02)--(0,0.14);\draw[line width=1.7pt] (0,0.05)--(0,0.11);
            \draw[line width=0.45pt] (0.13,0.0)--(0.13,0.12);\draw[line width=1.7pt] (0.13,0.03)--(0.13,0.09);},
}

%% file: fig_motivation.tex
%
\definecolor{cmWoad}    {HTML}{33506E}%
\definecolor{cmMurex}   {HTML}{6E3B5C}%
\definecolor{cmVerdi}   {HTML}{3D7A70}%
\definecolor{cmOrpi}    {HTML}{A87833}%
\definecolor{cmTerre}   {HTML}{5E7548}%
\definecolor{cmSinopia} {HTML}{9A4A34}%
\definecolor{cmBister}  {HTML}{33302B}%
\colorlet{sky}{cmWoad}%
\colorlet{amber}{cmOrpi}%
\colorlet{coral}{cmSinopia}%
\colorlet{sea}{cmVerdi}%
\colorlet{leaf}{cmTerre}%
\colorlet{lavender}{cmMurex}%
\colorlet{ink}{cmBister}%
\begin{tikzpicture}[x=1cm,y=1cm,
  every node/.style={font=\scriptsize},
  badge/.style={rounded corners=3pt, align=center, inner sep=3pt,
                font=\fontsize{6.6}{7.6}\bfseries\selectfont},
  lane/.style={font=\fontsize{7.6}{8.4}\bfseries\selectfont},
  note/.style={rounded corners=3pt, align=center, inner sep=2.6pt, draw=#1!55!black,
               fill=#1!11, font=\fontsize{6.2}{7.2}\selectfont},
  farr/.style={-{Stealth[length=1.7mm]}, line width=1.0pt, draw=#1}]

\def\x0{1.6}\def\W{2.68}
\foreach \i/\c/\cs/\t in {0/sky/cup/Growth, 1/amber/cup/Inflation, 2/coral/cdn/Stress,
                          3/sea/cv/Recovery, 4/leaf/cfl/Range}{
  \fill[\c!72, rounded corners=1.5pt] (\x0+\i*\W,7.50) rectangle (\x0+\i*\W+\W,7.93);
  \pic[white, line width=0.7pt] at (\x0+\i*\W+0.42,7.715) {\cs};
  \node[white, font=\fontsize{6.4}{6.8}\bfseries\selectfont] at (\x0+\i*\W+\W/2+0.15,7.715) {\t};
}
\node[font=\fontsize{6.6}{7}\itshape\selectfont, text=ink!70, anchor=east] at (\x0-0.15,7.715) {regime};
\node[draw=ink!45, fill=ink!06, rounded corners=1.6pt, inner sep=2.2pt,
      font=\fontsize{6.4}{6.8}\bfseries\itshape\selectfont, text=ink!65,
      anchor=west] at (1.0,6.92) {Illustrative scenario -- not experimental results};
\draw[-{Stealth[length=1.5mm]}, line width=0.9pt, ink!45] (15.05,7.715) -- (15.7,7.715);
\node[font=\fontsize{6.2}{6.6}\itshape\selectfont, text=ink!55] at (15.85,7.37) {time};

\foreach \i in {1,2,3,4}{
  \draw[ink!16, line width=0.7pt, dash pattern=on 2pt off 2pt] (\x0+\i*\W,6.90) -- (\x0+\i*\W,1.15);}

\def\Lv{5.60}\def\Lc{3.87}\def\Lm{2.15}
\node[lane, text=coral!55!black, rotate=90] at (0.55,\Lv) {Frozen agent};
\node[lane, text=amber!50!black, rotate=90] at (0.55,\Lc) {Fixed workflow};
\node[lane, text=leaf!50!black, rotate=90] at (0.55,\Lm) {\textsc{Cmas} (ours)};

\pic[coral!60!black, line width=0.75pt] at (1.85,\Lv) {robot};
\node[font=\fontsize{5.8}{6.2}\selectfont, text=coral!50!black] at (1.85,\Lv-0.46) {frozen};
\draw[farr=coral!75] (2.3,\Lv) -- (13.9,\Lv);
\node[note=coral, text width=2.1cm, anchor=south] at (4.6,\Lv+0.28) {stale harness $\cdot$ delayed feedback never reused};
\node[xnode, minimum size=11pt] (vx) at (\x0+2*\W,\Lv) {\ding{55}};
\node[font=\fontsize{6}{6.4}\bfseries\selectfont, text=coral!50!black, anchor=south] at (\x0+2*\W,\Lv+0.52) {shift missed, no update};
\pic[coral, line width=1.0pt] at (13.35,\Lv) {downarr};
\node[badge, draw=coral!65!black, fill=coral!15, text=coral!45!black, anchor=west] at (14.35,\Lv) {possible failure under shift};

\foreach \x/\y/\b/\bc in {%
  2.35/4.62/rtrend/sky, 2.95/4.62/rglobe/amber, 3.55/4.62/rcoin/leaf,
  2.35/4.06/rbubble/lavender, 2.95/4.06/rumbrella/coral, 3.55/4.06/rq/sea}{
  \pic[ink, line width=0.5pt, scale=0.6] at (\x,\y) {robot};
  \pic[\bc!65!black, line width=0.7pt, scale=0.82] at (\x+0.17,\y+0.19) {\b};
}
\node[font=\fontsize{5.6}{6}\bfseries\selectfont, text=ink!70, anchor=south west]
  at (2.05,4.92) {6 specialists: trend\,$\cdot$\,macro\,$\cdot$\,value\,$\cdot$\,sent.\,$\cdot$\,tail\,$\cdot$\,skeptic};
\foreach \cx in {5.62,8.30,10.98,13.66}{
  \foreach \dx/\dy in {-0.17/0.08,0/0.08,0.17/0.08,-0.17/-0.08,0/-0.08,0.17/-0.08}{
    \fill[ink!26] (\cx+\dx,\Lc-0.02+\dy) circle (0.03);}}
\draw[-{Stealth[length=1.6mm]}, line width=0.9pt, amber!70!black, dash pattern=on 2.4pt off 1.8pt]
  (4.3,\Lc) -- (14.0,\Lc);
\node[font=\fontsize{6.2}{6.8}\itshape\selectfont, text=amber!45!black] at (8.9,\Lc-0.42) {identical roster, every context};
\pic[amber!70!black, line width=0.7pt] at (11.35,\Lc+0.42) {coins};
\node[badge, draw=amber!60!black, fill=amber!15, text=amber!40!black] at (12.05,\Lc+0.42) {higher fixed inference cost};
\node[badge, draw=amber!60!black, fill=amber!15, text=amber!40!black, anchor=west] at (14.35,\Lc) {may remain adequate};

\foreach \x/\b/\bc in {%
  2.70/rtrend/sky, 3.20/rcoin/leaf,
  5.37/rglobe/amber, 5.87/rumbrella/coral,
  7.90/rglobe/amber, 8.30/rumbrella/coral, 8.70/rq/sea,
  10.73/rtrend/sky, 11.23/rbubble/lavender,
  13.41/rtrend/sky, 13.91/rq/sea}{
  \pic[ink, line width=0.5pt, scale=0.62] at (\x,\Lm) {robot};
  \pic[\bc!65!black, line width=0.72pt, scale=0.8] at (\x+0.16,\Lm+0.2) {\b};
}
\node[font=\fontsize{5.6}{6}\itshape\selectfont, text=leaf!45!black] at (2.95,\Lm-0.42) {lean};
\node[font=\fontsize{5.6}{6}\itshape\selectfont, text=leaf!45!black, anchor=west] at (1.05,\Lm+0.62) {continual harness adaptation};
\node[font=\fontsize{5.6}{6}\itshape\selectfont, text=coral!50!black] at (8.30,\Lm-0.42) {expand: +hedge};
\foreach \i/\lab in {1/,2/,3/,4/}{
  \pic[leaf!60!black, line width=0.7pt, scale=0.55] at (\x0+\i*\W,\Lm) {shield};}
\node[font=\fontsize{5.6}{6}\selectfont, text=leaf!45!black, anchor=north] at (\x0+1*\W,\Lm-0.28) {conservative gate};
\pic[leaf, line width=1.0pt] at (13.35,\Lm+0.48) {chk};
\node[badge, draw=leaf!60!black, fill=leaf!15, text=leaf!40!black, anchor=west] at (14.35,\Lm) {adapts under\\fixed risk controls};

\end{tikzpicture}

%% file: fig_overview.tex
%
%
\definecolor{cmWoad}    {HTML}{33506E}
\definecolor{cmMurex}   {HTML}{6E3B5C}
\definecolor{cmVerdi}   {HTML}{3D7A70}
\definecolor{cmOrpi}    {HTML}{A87833}
\definecolor{cmTerre}   {HTML}{5E7548}
\definecolor{cmSinopia} {HTML}{9A4A34}
\definecolor{cmBister}  {HTML}{33302B}
\definecolor{cmOat}     {HTML}{EFEAE0}
\definecolor{cmStone}   {HTML}{E7E4DD}
\definecolor{cmParch}   {HTML}{F4EEE0}

\begin{tikzpicture}[x=1cm,y=1cm,
  every node/.style={font=\scriptsize},
  band/.style={rounded corners=5pt, draw=cmBister!25, line width=0.7pt},
  bandttl/.style={anchor=west, font=\fontsize{7.4}{8.2}\sffamily\selectfont},
  chip/.style={rounded corners=3pt, draw=#1!92!black, fill=#1!13,
               line width=0.6pt, align=center, inner xsep=3pt, inner ysep=3.4pt,
               font=\fontsize{7.0}{8.4}\selectfont, text=cmBister!85},
  locked/.style={rounded corners=3pt, draw=cmBister!70, fill=cmBister!88,
                 line width=0.7pt, align=center, inner xsep=4pt, inner ysep=4pt,
                 font=\fontsize{7.0}{8.4}\selectfont, text=white},
  flow/.style={-{Stealth[length=1.6mm,width=1.3mm]}, line width=0.9pt, draw=cmBister!38},
  hair/.style={-{Stealth[length=1.2mm,width=1.0mm]}, line width=0.55pt, draw=cmBister!32},
  ret/.style={-{Stealth[length=1.9mm,width=1.5mm]}, line width=1.15pt, draw=cmTerre!52!black},
  ev/.style={-{Stealth[length=1.9mm,width=1.5mm]}, line width=1.15pt, draw=cmWoad!55!black},
  micro/.style={font=\fontsize{6.4}{7.2}\sffamily\selectfont, text=cmBister!48},
  num/.style={circle, fill=cmBister!80, text=white, inner sep=0pt,
              minimum size=3.4mm, font=\fontsize{6.2}{6.2}\bfseries\sffamily\selectfont}]

\hyphenpenalty=10000 \exhyphenpenalty=10000 \tolerance=1000

\fill[cmOat!75, rounded corners=5pt] (4.55,3.62) rectangle (19.75,6.72);
\draw[band]                            (4.55,3.62) rectangle (19.75,6.72);
\node[bandttl] at (4.82,6.36)
  {\textbf{\textcolor{cmBister!62}{OUTER LOOP}}\;\;\textcolor{cmBister!42}{monthly $\cdot$
   propose, screen and promote one bounded architecture update}};

\node[chip=cmWoad,      minimum height=0.95cm, text width=2.05cm] (o1) at ( 6.05,4.86) {\textbf{retrieve}\\parents from $\mathcal{D}_k$};
\node[chip=cmMurex, minimum height=0.95cm, text width=2.05cm] (o2) at ( 8.55,4.86) {\textbf{typed diff}\\compile-checked};
\node[chip=cmVerdi,      minimum height=0.95cm, text width=2.05cm] (o3) at (11.05,4.86) {\textbf{paired replay}\\shared seeds};
\node[chip=cmOrpi,    minimum height=0.95cm, text width=2.05cm] (o4) at (13.55,4.86) {\textbf{credit} $\phi_i$\\per component};
\node[chip=cmTerre,     minimum height=0.95cm, text width=2.05cm] (o5) at (16.05,4.86) {\textbf{gate}\\$\Score_k(c)>\epsilon$};
\node[chip=cmSinopia,    minimum height=0.95cm, text width=2.05cm] (o6) at (18.55,4.86) {\textbf{canary}\\$+$ rollback};

\foreach \a/\b in {o1/o2,o2/o3,o3/o4,o4/o5,o5/o6} {\draw[flow] (\a) -- (\b);}
\foreach \n/\s in {1/o1,2/o2,3/o3,4/o4,5/o5,6/o6}
  {\node[num] at ($(\s.north west)+(0.11,-0.03)$) {\n};}

\node[micro, anchor=north] at ($(o3.south)+(0,-0.07)$) {candidate vs.\ incumbent};
\node[micro, anchor=north] at ($(o5.south)+(0,-0.07)$) {empirically calibrated};
\node[micro, anchor=north] at ($(o6.south)+(0,-0.07)$) {reversible};

\fill[cmStone!60, rounded corners=5pt] (4.55,0.18) rectangle (19.75,3.06);
\draw[band]                             (4.55,0.18) rectangle (19.75,3.06);
\node[bandttl] at (4.82,2.76)
  {\textbf{\textcolor{cmBister!62}{INNER LOOP}}\;\;\textcolor{cmBister!42}{weekly $\cdot$
   run the deployed harness $h_k$}};

\draw[dashed, cmBister!32, line width=0.6pt, rounded corners=3pt]
      (4.80,0.78) rectangle (13.05,2.44);
\node[micro, anchor=south east, text=cmBister!42] at (13.01,2.48) {LLM-EDITABLE};

\node[chip=cmWoad,      minimum height=0.58cm, text width=1.10cm] (w1) at ( 5.66,2.00) {trend};
\node[chip=cmOrpi,    minimum height=0.58cm, text width=1.10cm] (w2) at ( 7.02,2.00) {macro};
\node[chip=cmSinopia,    minimum height=0.58cm, text width=1.10cm] (w3) at ( 8.38,2.00) {tail-risk};
\node[chip=cmMurex, minimum height=0.58cm, text width=1.10cm] (w4) at ( 9.74,2.00) {skeptic};
\node[chip=cmBister,      minimum height=0.58cm, text width=1.52cm] (agg) at (11.72,2.00) {aggregate $g$};

\coordinate (rail) at ($(w1.south)+(0,-0.34)$);
\coordinate (join) at ($(w4.east)!0.5!(agg.west)$);   
\foreach \w in {w1,w2,w3,w4}
  {\draw[cmBister!32, line width=0.55pt] (\w.south) -- (\w.south |- rail);}
\draw[cmBister!32, line width=0.55pt] (w1.south |- rail) -- (join |- rail);
\draw[hair, rounded corners=2.5pt] (join |- rail) -- (join) -- (agg.west);

\node[micro, anchor=north, inner sep=1pt] at ($(w1.south)!0.5!(agg.south) + (0,-0.42)$)
  {typed outputs $o_{i,t}$: scores $\cdot$ confidence $\cdot$ evidence $\cdot$ risk flags};

\node[locked, minimum height=0.95cm, text width=3.30cm] (lock) at (15.30,2.00)
  {\hspace{0.32cm}\textbf{risk projection} $\Pi_{\mathcal{C}_t}$\\[0.6pt]
   \hspace{0.32cm}authorization $\cdot$ accounting\\[0.6pt]
   \hspace{0.32cm}$\cdot$ data cutoffs};
\pic[white, line width=0.85pt] at ($(lock.west)+(0.32,0.02)$) {lock};
\node[micro, anchor=south, text=cmBister!58] at ($(lock.north)+(0,0.05)$)
  {IMMUTABLE $\cdot$ the proposer cannot edit this};

\node[chip=cmTerre, minimum height=0.58cm, text width=1.24cm] (dec) at (18.60,2.00) {decision $w_t$};

\draw[flow] (agg) -- (lock);
\draw[flow] (lock) -- (dec);

\node[chip=cmVerdi, minimum height=0.44cm, text width=7.10cm] (tr) at (9.30,0.44)
  {trace $\tau_t$: tool calls $\cdot$ cost $\cdot$ violations $\cdot$ delayed outcome};
\draw[hair] (dec.south) |- (tr.east);

\fill[cmParch!70, rounded corners=5pt] (0.18,0.18) rectangle (4.05,6.72);
\draw[band, draw=cmOrpi!45!black]         (0.18,0.18) rectangle (4.05,6.72);
\node[bandttl, text=cmOrpi!32!black] at (0.44,6.36) {\textbf{ARCHIVE $\mathcal{D}_k$}};
\node[micro, text=cmOrpi!28!black, text width=3.35cm, align=left, anchor=north west]
   at (0.44,6.06) {operational memory of prior workflow behaviour};

\foreach \y/\txt in {%
  4.86/{regime $z_j$ $\cdot$ architecture $h_j$},
  4.02/{trace $\cdot$ utility $\hat J_j$},
  3.18/{credit $\hat\phi_j$ $\cdot$ failure label}}{
  \node[rounded corners=2.5pt, draw=cmOrpi!50!black, fill=white, line width=0.55pt,
        minimum height=0.62cm, text width=2.95cm, align=center, inner xsep=2pt,
        font=\fontsize{6.6}{7.4}\selectfont, text=cmBister!78] at (2.11,\y) {\txt};}

\node[micro, text width=3.35cm, align=left, anchor=north west, text=cmOrpi!28!black]
  at (0.44,2.58)
  {conditions the next proposal; the foundation-model weights are never updated};

\draw[ev] (tr.west) -- (4.11,0.44);
\node[micro, anchor=south, text=cmWoad!45!black] at (4.34,0.50) {log};

\draw[ev] (4.11,4.86) -- (o1.west);

\draw[ret] (o6.south) -- (18.55,3.40) -- (18.55,3.14);
\node[micro, anchor=east, text=cmTerre!38!black] at (18.40,3.28)
  {promote $h_{k+1}$, or retain $h_k$};

\pgfresetboundingbox
\path[use as bounding box] (0.165,0.165) rectangle (19.765,6.735);
\end{tikzpicture}%

%% file: fig_bench.tex
\begin{tikzpicture}[x=1cm,y=1cm,
  every node/.style={font=\scriptsize},
  track/.style={rounded corners=6pt, draw=#1!60!black, fill=#1!7, line width=0.9pt},
  ttl/.style={font=\fontsize{7.4}{8.2}\bfseries\selectfont, text=#1!45!black},
  desc/.style={font=\fontsize{6.8}{7.6}\selectfont, text=ink!80, align=left},
  tedge/.style={draw=ink!25, line width=0.8pt, dash pattern=on 2.4pt off 2pt},
  hubarr/.style={-{Stealth[length=2mm]}, line width=1.0pt, draw=#1!55!black}]

\def\Ax{2.75} \def\Ay{5.35}
\def\Bx{11.45} \def\By{5.35}
\def\Cx{7.1}  \def\Cy{0.75}
\def\Hx{7.1}  \def\Hy{3.75}

\draw[tedge] (\Ax+0.2,\Ay-0.55) -- (\Bx-0.2,\By-0.55);
\draw[tedge] (\Ax-0.1,\Ay-0.75) -- (\Cx-1.7,\Cy+0.55);
\draw[tedge] (\Bx+0.1,\By-0.75) -- (\Cx+1.7,\Cy+0.55);

\node[track=sky, minimum width=4.3cm, minimum height=1.85cm] (A) at (\Ax,\Ay) {};
\pic[sky!55!black, line width=0.9pt] at ($(A.north west)+(0.42,-0.5)$) {flask};
\pic[ink, line width=0.7pt] at ($(A.north west)+(0.28,-1.16)$) {hmm};
\node[ttl=sky, anchor=north west] at ($(A.north west)+(0.82,-0.16)$) {Track A \textperiodcentered\ controlled simulator};
\node[desc, anchor=north west, text width=3.4cm] at ($(A.north west)+(0.82,-0.5)$)
  {\emph{simulator ground truth.} Hidden regimes $s_t$, competence $H_{i,s}$, surrogate-optimal architecture};
\node[font=\fontsize{6.6}{7.3}\bfseries\selectfont, text=sky!45!black, anchor=south west]
  at ($(A.south west)+(0.82,0.1)$) {regret \textperiodcentered\ selection acc \textperiodcentered\ credit fidelity};

\node[track=sea, minimum width=4.3cm, minimum height=1.85cm] (B) at (\Bx,\By) {};
\pic[sea!55!black, line width=0.9pt] at ($(B.north west)+(0.42,-0.5)$) {candles};
\pic[leaf!60!black, line width=0.8pt] at ($(B.north west)+(0.30,-1.18)$) {shield};
\node[ttl=sea, anchor=north west] at ($(B.north west)+(0.82,-0.16)$) {Track B \textperiodcentered\ market replay};
\node[desc, anchor=north west, text width=3.4cm] at ($(B.north west)+(0.82,-0.5)$)
  {\emph{date-controlled.} Publication lags + counterfactual resampling; no tickers/dates};
\node[font=\fontsize{6.6}{7.3}\bfseries\selectfont, text=sea!45!black, anchor=south west]
  at ($(B.south west)+(0.82,0.1)$) {CEQ \textperiodcentered\ Sharpe \textperiodcentered\ factor-$\alpha$ \textperiodcentered\ drawdown};

\node[track=coral, minimum width=4.7cm, minimum height=1.7cm] (C) at (\Cx,\Cy) {};
\pic[coral!60!black, line width=0.9pt] at ($(C.north west)+(0.42,-0.62)$) {bug};
\pic[leaf!60!black, line width=0.9pt] at ($(C.north west)+(0.98,-0.62)$) {shield};
\node[ttl=coral, anchor=north west] at ($(C.north west)+(1.4,-0.18)$) {Track C \textperiodcentered\ governance suite};
\node[desc, anchor=north west, text width=3.0cm] at ($(C.north west)+(1.4,-0.52)$)
  {\emph{budgeted adversary} corrupts workers and data};
\node[font=\fontsize{6.6}{7.3}\bfseries\selectfont, text=coral!45!black, anchor=south west]
  at ($(C.south west)+(1.4,0.08)$) {detect \textperiodcentered\ isolate \textperiodcentered\ rollback \textperiodcentered\ recover};

\node[circle, draw=ink!45, fill=softgray!55, line width=1.0pt, minimum size=1.85cm] (H) at (\Hx,\Hy) {};
\pic[ink, line width=1.0pt] at ($(H.center)+(0,0.22)$) {gauge};
\node[font=\fontsize{7.0}{7.8}\bfseries\selectfont, text=ink!75, align=center] at ($(H.center)+(0,-0.22)$)
  {system-level\\ adaptation metrics};

\draw[hubarr=sky] (A.south) to[bend right=8] ($(H.north west)+(-0.05,0.05)$);
\draw[hubarr=sea] (B.south) to[bend left=8] ($(H.north east)+(0.05,0.05)$);
\draw[hubarr=coral] (C.north) -- (H.south);

\node[draw=ink!30, fill=white, rounded corners=3pt, inner sep=3pt,
      font=\fontsize{6.6}{7.3}\selectfont, text=ink!80, align=center, text width=8.4cm]
  at (\Hx,\Hy-1.42)
  {regret \& delay \ \textperiodcentered\ \ OOD utility \& transfer \ \textperiodcentered\ \ credit fidelity
   \ \textperiodcentered\ \ false promotion \& rollback \ \textperiodcentered\ \ sparsity \ \textperiodcentered\ \ cost accounting};

\node[font=\fontsize{6.8}{7.6}\itshape\selectfont, text=ink!60, align=center]
  at (\Hx,-0.55)
  {one backtest cannot certify a continually adapting agent: \textbf{controlled ground truth} (A), \textbf{external validity} (B), and \textbf{adversarial robustness} (C) are triangulated};
\end{tikzpicture}